\documentclass[nohyperref]{article}

\usepackage{microtype}
\usepackage{graphicx}
\usepackage{subfigure}
\usepackage{booktabs} 

\usepackage{hyperref}

\usepackage[accepted]{icml2023}

\usepackage{amsmath}
\allowdisplaybreaks[4]
\usepackage{amssymb}
\usepackage{mathtools}
\usepackage{amsthm}
\usepackage{dsfont}

\usepackage{amsmath,amsfonts,bm}

\newtheorem{theorem}{Theorem}[section]
\newtheorem{lemma}{Lemma}[section]
\newtheorem{definition}{Definition}[section]
\newtheorem{assumption}{Assumption}[section]
\newtheorem{proposition}{Proposition}[section]
\newtheorem{fact}{Fact}[section]

\def\eqref#1{equation~\ref{#1}}

\def\1{\bm{1}}

\def\eps{{\epsilon}}

\DeclareMathAlphabet{\mathsfit}{\encodingdefault}{\sfdefault}{m}{sl}
\SetMathAlphabet{\mathsfit}{bold}{\encodingdefault}{\sfdefault}{bx}{n}

\def\gA{{\mathcal{A}}}

\def\gE{{\mathcal{E}}}
\def\gF{{\mathcal{F}}}

\def\gM{{\mathcal{M}}}

\def\gP{{\mathcal{P}}}

\def\gS{{\mathcal{S}}}

\def\sB{{\mathbb{B}}}

\def\sP{{\mathbb{P}}}

\def\sR{{\mathbb{R}}}

\newcommand{\E}{\mathbb{E}}

\newcommand{\R}{\mathbb{R}}

\DeclareMathOperator*{\argmax}{arg\,max}
\DeclareMathOperator*{\argmin}{arg\,min}

\newcommand{\spr}{s^{\prime}}

\newcommand{\maxb}{\max_{\beta_{h,i}\in [\underline{\beta}, \overline{\beta}]}}
\newcommand{\epsV}{V^{\prime}}
\newcommand{\epsN}{\mathcal{N}_{\epsilon}}
\newcommand{\betaeps}{\mathcal{N}_{\epsilon}(\underline{\beta}, \overline{\beta})}

\newcommand{\robvalue}{V^{\pi, \operatorname{rob}}_h(s)}
\newcommand{\robQ}{Q^{\pi, \operatorname{rob}}_h(s,a)}
\newcommand{\robR}[1]{R^{\operatorname{rob}}(#1, \gP)}
\newcommand{\expbeta}{e^{2H/\underline{\beta}}\cdot \frac{4H}{\beta^2}\epsilon}
\newcommand{\epsbeta}{2\expbeta}
\newcommand{\epsVValue}{e^{H/\underline{\beta}}\frac{2\epsilon}{\beta}}
\newcommand{\epsbetas}{\frac{4N^2(H-h)^2\epsilon^2}{\lambda\underline{\beta}^4}e^{2H/\underline{\beta}}}
\newcommand{\epsVs}{\frac{4N^2\epsilon^2}{\lambda\underline{\beta}^2}e^{2H/\underline{\beta}}}

\newcommand{\zetaI}{\log(2N + 16Nd^{3/2}H^2e^{H/\underline{\beta}})}
\newcommand{\zetaII}{\log(\frac{2dNH^3}{\delta\rho})}
\newcommand{\zetaIII}{\log(2N+32N^2H^3d^{5/2}\zeta e^{2H/\beta})}
\newcommand{\gammah}{c_1\underline{\beta}(e^{\frac{H-h}{\underline{\beta}}}\!-\!1) d\zeta_3^{1/2}+ c_2\underline{\beta}^{1/2}(e^{\frac{H-h}{\underline{\beta}}}\!-\!1)H^{1/2}\zeta_2^{1/2}}

\usepackage[capitalize,noabbrev]{cleveref}

\theoremstyle{plain}
\theoremstyle{definition}
\theoremstyle{remark}

\usepackage[textsize=tiny]{todonotes}

\icmltitlerunning{Distributionally Robust Offline RL with Linear Function Approximation}

\begin{document}

\twocolumn[
\icmltitle{Distributionally Robust Offline Reinforcement Learning \\
with Linear Function Approximation}



\icmlsetsymbol{equal}{*}

\begin{icmlauthorlist}
    \icmlauthor{Xiaoteng Ma}{equal,thu}
    \icmlauthor{Zhipeng Liang}{equal,hkust}
    \icmlauthor{Jose Blanchet}{stanford}
    \icmlauthor{Mingwen Liu}{sysu}\\
    \icmlauthor{Li Xia}{sysu}
    \icmlauthor{Jiheng Zhang}{hkust}
    \icmlauthor{Qianchuan Zhao}{thu}
    \icmlauthor{Zhengyuan Zhou}{nyu}
\end{icmlauthorlist}

\icmlaffiliation{hkust}{Hong Kong University of Science and Technology}
\icmlaffiliation{thu}{Tsinghua University}
\icmlaffiliation{stanford}{Stanford University}
\icmlaffiliation{nyu}{New York University}
\icmlaffiliation{sysu}{Sun Yat-sen University}

\icmlcorrespondingauthor{Zhipeng Liang}{zliangao@connect.ust.hk}
\icmlcorrespondingauthor{Xiaoteng Ma}{ma-xt17@mails.tsinghua.edu.cn}

\icmlkeywords{Machine Learning, ICML}

\vskip 0.3in
]



\printAffiliationsAndNotice{\icmlEqualContribution} 

\begin{abstract}
    Among the reasons hindering reinforcement learning (RL) applications to real-world problems, two factors are critical: limited data and the mismatch between the testing environment (real environment in which the policy is deployed) and the training environment (e.g., a simulator). This paper attempts to address these issues simultaneously with \emph{distributionally robust offline RL}, where we learn a distributionally robust policy using historical data obtained from the source environment by optimizing against a worst-case perturbation thereof.
    In particular, we move beyond tabular settings and consider linear function approximation. More specifically, we consider two settings, one where the dataset is well-explored and the other where the dataset has sufficient coverage of the optimal policy.
    We propose two algorithms~-- one for each of the two settings~-- that achieve error bounds $\tilde{O}(d^{1/2}/N^{1/2})$ and $\tilde{O}(d^{3/2}/N^{1/2})$ respectively, where $d$ is the dimension in the linear function approximation and $N$ is the number of trajectories in the dataset.
    To the best of our knowledge, they provide the first non-asymptotic results of the sample complexity in this setting. 
    Diverse experiments are conducted to demonstrate our theoretical findings, showing the superiority of our algorithm against the non-robust one.
\end{abstract}
    
\section{Introduction}Unlike the data-driven methods in supervised learning, reinforcement learning (RL) algorithms learn a near-optimal policy by actively interacting with the environment, typically involving online trial-and-error to improve the policy.
However, data collection could be expensive, and the collected data may be limited in many real-world problems, making online interaction with the environment limited or even prohibited.
To address the limitation of online RL, offline Reinforcement Learning (offline RL or batch RL)~\citep{lange2012batch, levine2020offline}, a surging topic in the RL, focuses on policy learning with only access to some logged datasets and expert demonstrations.
Benefiting from its non-dependence on further interaction with the environment, offline RL is appealing for a wide range of applications, including autonomous driving~\citep{yu2018bdd100k, yurtsever2020survey, shi2021offline}, healthcare~\citep{gottesman2019guidelines,yu2021reinforcement,tang2021model} and robotics~\citep{siegel2020keep,zhou2021plas,rafailov2021offline}.


Despite the advances achieved in the rich literature~\citep{yu2020mopo,kumar2020conservative,yang2021believe,an2021uncertainty,cheng2022adversarially}, offline RL has an implicit but questionable assumption: the test environment is the same as the training one.
This assumption may lead to the ineffectiveness of the offline RL in uncertain environments as the optimal policy of an MDP may be very sensitive to the transition probabilities~\citep{mannor2004bias, el2005robust, simester2006dynamic}.
Many domains, such as financial trading and robotics, may prefer a robust policy that remains effective in shifting distributions from the one in the training environment.
Thus, robust MDPs and corresponding algorithms have been proposed to address this issue~\citep{satia1973markovian, nilim2005robust, iyengar2005robust, wiesemann2013robust, lim2013reinforcement,ho2021partial,goyal2022robust}. 
Recently, a stream of works~\citep{zhou2021finite,yang2021towards,shi2022distributionally,panaganti2022robust} studies robust RL in the offline setting, showing the promise of learning robust policies with offline datasets.


In this paper, we aim to theoretically understand linear function approximation as an important component in distributionally robust offline RL.
It is well-known that the curse of dimensionality prohibits the employment of DP-based algorithms for problems with large state-action spaces.
Approximate dynamic programming (ADP) \citep{bertsekas2008approximate}, which approximates the value functions with some basis functions, arouses the most popular solution to the curse of dimensionality.
Among all the choices of approximation, linear function approximation \citep{bertsekas1995neuro, schweitzer1985generalized}, which uses a linear combination of features to approximate the value function, is the most studied approach and serves as a cornerstone in the path toward real-world problems.

Compared with the nominal RL algorithms, designing a distributionally robust RL algorithm with linear function approximation is highly nontrivial.
An immediate attempt in this topic \citep{tamar2013scaling} projects the robust value function onto a lower dimensional subspace by means of linear function approximation.
Although they can prove their robust projected value iteration can converge to a fixed point, as pointed out by our motivating example in Section~\ref{sec:motivation}, the linear projection may not suit the non-linearity of the robust Bellman operator and may consequently lead to wrong decisions. 
Moreover, to the best of our knowledge, there are no provably distributionally robust linear RL algorithms with non-asymptotic suboptimality results.

In this paper, we make an attempt to answer the question:

\emph{Is it possible to design a provable sample-efficient algorithm using linear function approximation for the distributionally robust offline RL?}

In particular, our contributions are four folds:
\begin{enumerate}
    \item We formulate a provable distributionally robust offline RL model based on linear function approximation and design the first sample-efficient \textbf{D}istributionally \textbf{R}obust \textbf{V}alue \textbf{I}teration with \textbf{L}inear function approximation (DRVI-L) algorithm for well-explored datasets.
    \item We prove a state-action space independent suboptimality for our DRVI-L algorithm with a novel value shift technique to alleviate the magnification of the estimation error in the distributionally robust optimization nature.
    The suboptimality can almost recover to the same dependence on $d$ and $N$ of from the non-robust counterpart~\cite{yin2022nearoptimal}.
    \item We extend our algorithm by designing the \textbf{P}essimistic \textbf{D}istributionally \textbf{R}obust \textbf{V}alue \textbf{I}teration with \textbf{L}inear function approximation (PDRVI-L) algorithm, a pessimistic variant with our DRVI-L, and prove a sample-efficient bound beyond the well-explored condition.
    \item We provide theoretical guarantees for our two algorithms even when the MDP transition is not perfectly linear and conduct experiments to demonstrate the balance struck by our linear function approximation algorithm between optimality and computational efficiency.
\end{enumerate}

\subsection{Related Work}
\textbf{Offline RL:}
Recent research interests arouse to design offline RL algorithms with fewer dataset requirements based on a shared intuition called pessimism, i.e., the agent can act conservatively in the face of state-action pairs that the dataset has not covered.
Empirical evidence has emerged~\citep{fujimoto2019off, wu2019behavior, kumar2019stabilizing, fujimoto2021minimalist, kumar2020conservative, kostrikov2021offline, pmlr-v139-wu21i, wang2018exponentially, chen2020bail, yang2021believe, kostrikov2022offline}. 
\citet{jin2021pessimism} prove that a pessimistic variant of the value iteration algorithm can achieve sample-efficient suboptimality under a mild data coverage assumption.
\citet{xie2021bellman} introduce the notion of Bellman consistent pessimism to design a general function approximation algorithm. 
\citet{rashidinejad2021bridging} design the lower confidence bound algorithm utilizing pessimism in the face of uncertainty and show it is almost adaptively optimal in MDPs.

\textbf{Linear Function Approximation:} 
Research interests on the provable efficient RL under the linear model representations have emerged in recent years.
\citet{yang2019sample} propose a parametric $Q$-learning algorithm to find an approximate-optimal policy with access to a generative model.
\citet{zanette2021provable} considers the Linear Bellman Complete model and designs the efficient actor-critic algorithm that achieves improvement in dependence on $d$.
\citet{yin2022nearoptimal} designs the variance-aware pessimistic value iteration to improve the suboptimality bounds over the best-known existing results.
On the other hand, \citet{wang2020statistical,zanette2021exponential} prove the statistical hardness of offline RL with linear representations by proving that the sample sizes could be exponential in the problem horizon for the value estimation task of any policy.


\textbf{Robust MDP and RL:}
The robust optimization approach has been used to address the parameters uncertainty in MDPs first by \citet{satia1973markovian} and later by \citet{xu2010distributionally, iyengar2005robust,nilim2005robust, wiesemann2013robust, kaufman2013robust, ho2018fast, ho2021partial, wiesemann2013robust}.
Although flourishing in the supervised learning \citep{namkoong2017variance,bertsimas2018data,duchi2021learning,duchi2021statistics}, 
few works consider computing the optimal robust policy for RL.
For online RL, a line of work has considered learning the optimal MDP policy under worst-case perturbations of the observation or environmental dynamics \citep{rajeswaran2016epopt,pattanaik2017robust, huang2017adversarial,pinto2017robust,zhang2020robust}.
For offline RL,~\citet{zhou2021finite} studies the distributionally robust policy with the offline dataset, where they focus on the KL ambiguity set and $(s,a)$-rectangular assumption and develop a value iteration algorithm.
\citet{yang2021towards} improve the results in \citet{zhou2021finite} and extend the algorithms to other uncertainty sets. 
However, current theoretical advances mainly focus on tabular settings.  

Among the previous work, one of the closest works to ours is~\citet{tamar2013scaling}, which develops a robust ADP method based on a projected Bellman equation.
Based on this,~\citet{badrinath2021robust} address the model-free robust RL with large state spaces by the proposed robust least squares policy iteration algorithm.
While both provide the convergence guarantee for their algorithm, as shown in Section~\ref{sec:motivation}, the projection into the linear space may lead to the wrong decision.
The other closed work to ours is~\citet{goyal2022robust}, which considers a more general assumption for the ambiguity set, called $d$-rectangular \footnote{\citet{goyal2022robust} call it $r$-rectangular.} 
for MDPs with low dimensional linear representation.
They mainly focus on the optimal policy structure for robust MDPs and the computational cost given the ambiguity set. In contrast, we study the offline RL setting and focus on the linear function approximation with a provable finite-sample guarantee for the suboptimality.


\section{Preliminary}\subsection{MDP structure and Notations}
Consider an episode MDP $(\gS, \gA, H, \mu, P, r)$ where $\gS$ and $\gA$ are finite state and action spaces with cardinalities $S$ and $A$.
$P=\{P_h\}_{h=1}^H$ are state transition probability measures and $r=\{r_h\}_{h=1}^H$ are the reward functions, respectively. 
Without loss of generality, we assume that $r$ is deterministic and bounded in $[0, 1]$. 
A (Markovian) policy $\pi=\{\pi_h\}_{h=1}^H$ maps, for each period state-action pair $(s,a)$ to a probability distribution over the set of actions $\gA$ and induce a random trajectory $s_1,a_1,r_1,\cdots,s_H, a_H, r_H, s_{H+1}$ with $s_1\sim \mu$, $a_h\sim \pi(\cdot\lvert s_h)$and $s_{h+1}\sim P_h(\cdot\lvert s_h, a_h)$  for $h\in [H]$  for some initial state distribution $\mu$. 
For any policy $\pi$ and any stage $h\in [H]$, the value function $V_h^{\pi}:\gS\rightarrow \R$, the action-value function $Q:\gS\times \gA \rightarrow \R$, the expected return $R(\pi, P)$ are defined as  $ V^{\pi}_h(s) \coloneqq \E_{P}^{\pi}[\sum_{h=1}^{H} r_{h}(s_h,a_h)\lvert s_h = s]$, $Q^{\pi}_h(s,a) \coloneqq \E_{P}^{\pi}[\sum_{h=1}^{H} r_{h}(s_h,a_h)\lvert s_h = s, a_h = a]$, and $ R(\pi, P) \coloneqq \E_{s\sim \mu}[V^{\pi}_1(s)]$.
For any function $Q$ and any policy $\pi$, we denote $\langle Q(s,\cdot), \pi(\cdot \lvert s)\rangle_{\gA} = \sum_{a\in \gA}Q(s,a)\pi(a\lvert s)$.
For two non-negative sequences $\{a_n\}$ and $\{b_n\}$, we denote $\{a_n\} = O(\{b_n\})$ if $\lim\sup_{n\rightarrow \infty}a_n/b_n<\infty$. We also use $\tilde{O}(\cdot)$ to  denote the respective meaning within multiplicative logarithmic factors in $N$, $d$, $H$ and $\delta$.
We denote the Kullback-Leibler (KL) divergence between two discrete probability distributions $P$ and $Q$ over state space as $D_{\operatorname{KL}}(P\lVert Q)=\sum_{s\in \gS}P(s)\log(\frac{P(s)}{Q(s)})$.


\subsection{Distributionally Robust Offline RL}
Before we present the distributionally robust RL setting, we first introduce the distributionally robust MDP.
Different from its non-robust counterpart, the true transition probability $P$ of distributionally robust MDP is not exactly known but lies within a so-called ambiguity set $\gP$. 
The notion of distributionally robust (DR) emphasizes that the ambiguity set is induced by a distribution distance measure, such as KL divergence.
The return of any given policy is the worst-case return induced by the transition model over the ambiguity set.
We define the DR value function, action-value function and expected return as $\robvalue = \inf_{P\in \gP}V^{\pi}_h(s)$, $\robQ = \inf_{P\in \gP}Q^{\pi}_h(s,a)$ and $\robR{\pi} = \inf_{P\in \gP} R(\pi, P)$.
The optimal DR expected return is defined as $\robR{\pi^*} \coloneqq \sup_{\pi\in \Pi}\robR{\pi} $ over all Markovian policies. In the sequel, we omit the superscript ``rob''.
In fact, by the work of~\citet{goyal2022robust}, we can restrict to the deterministic policy class to achieve the optimal DR expected return.
The performance metric for any given policy $\pi$ is the so-called suboptimality, which is defined as  
\begin{align*}
    \operatorname{SubOpt}(\pi; \gP) &= R({\pi^{*}},\gP) - R(\pi, \gP).
\end{align*}



As the transition models and the policies are sequences corresponding to all horizons in the episode MDP,
following~\citet{iyengar2005robust}, we assume that $\gP$ can be decomposed as the product of the ambiguity sets in each horizon, i.e., $\gP=\prod_{h=1}^H \gP_h$. 
For stage $h$, each transition model $P_h$, lies within the ambiguity set $\gP_h$.
\subsection{Linear MDP}
Our main task in this paper is to use the linear function to compute the policy from the data with the possible lowest suboptimality.
Given the feature map $\phi:\gS\times \gA\rightarrow\sR^{d}$, for each horizon $h\in[H]$, we parameterize the Q-function, value function and the induced policy using $\nu_h\in \R^d$ by 
\begin{align}
\label{eq:parameter}
    Q_{\nu_h}(s,a)&\coloneqq \phi(s,a)^{\top}\nu_h,\\  V_{\nu_h}(s)&\coloneqq \max_{a\in \gA}Q_{\nu_h}(s,a),\\
    \pi_{\nu_h}(s)&\coloneqq \arg\max_{a\in \gA} Q_{\nu_h}(s,a).
\end{align}
Various assumptions on the MDP have been proposed to study the linear function approximation~\citep{jiang2017contextual,yang2019sample,jin2020provably,modi2020sample, zanette2020learning, wang2021optimism}.
In particular, we consider the MDP with a soft state aggregation of $d$ factors, i.e., each state can be represented using a known feature map $\phi:\gS\times \gA \rightarrow \R^d$ over the $d$ factors $\psi:\gS\rightarrow \R^d$.
Such a assumption has been widely adopted in the literature~\citep{singh1994reinforcement, duan2019state, zhang2019spectral,zanette2021provable}. 
Moreover, we also assume the reward functions admit linear structure w.r.t. $\phi$, following the linear MDP protocol from \citet{jin2021pessimism,jin2020provably}. 
Formally, we have the following definition for the soft state aggregation.
\begin{definition}[Soft State Aggregation MDP]
    \label{def:linear}
Consider an episode MDP instance $M=(\gS,\gA, H, P, r)$ and a feature map $\phi: \gS\times \gA\rightarrow \R^{d}$. We say the transition model $P$ admit a soft state aggregation w.r.t. $\phi$ (denoted as $P\in \operatorname{Span}(\phi)$) if for every $s\in \gS, a\in \gA, \spr\in \gS$ and every $h\in[H]$, we have  
$$P_h(\spr\lvert s,a) = \phi(s,a)^{\top} \psi_{h}(\spr),$$
for some factors $\psi_h :\gS \rightarrow \R^{d}$. 
Moreover, $\psi$ satisfies,
\begin{align*}
    \int_s \psi_{h, i}(s) ds &=1, \forall i\in[d], h\in[H].
\end{align*}
We say the reward functions $r$ admit a linear representation w.r.t. $\phi$ (denoted as $r\in \operatorname{Span}(\phi)$) if for all $s\in \gS$ and $a\in \gA$ and $h\in [H]$, there exists $\theta_h\in \R^d$ satisfying $\lVert\theta_h \rVert\le \sqrt{d}$ and 
$r_h(s,a) = \phi(s,a)^{\top}\theta_h$.
\end{definition}

\section{Motivating Example}\label{sec:motivation}
\begin{figure}
    \centering
    \includegraphics[width=6cm]{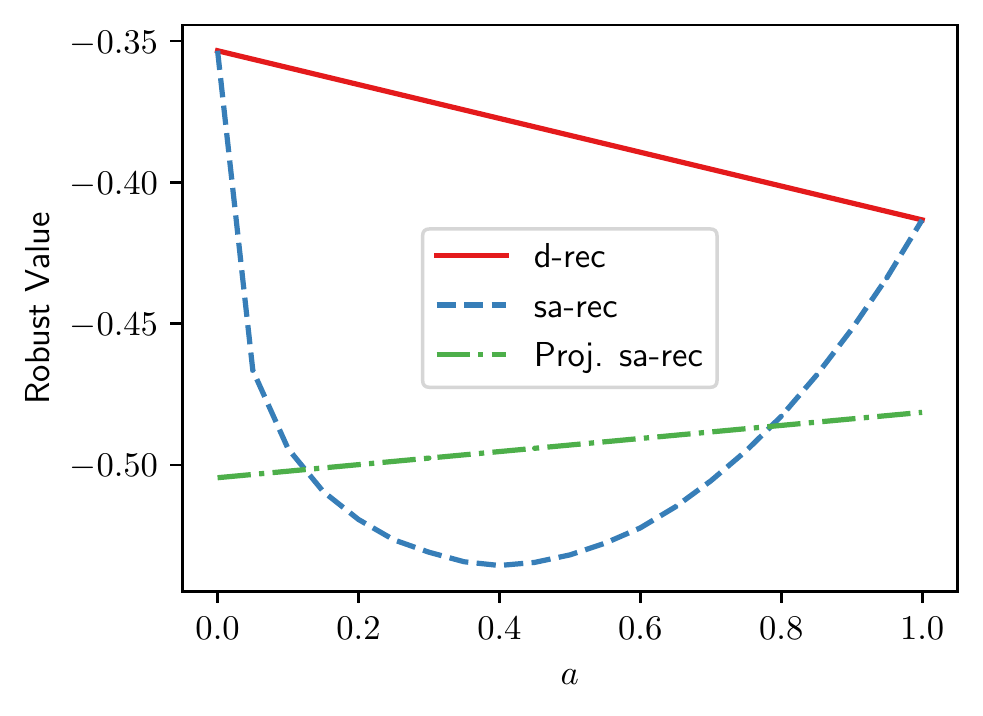}
    \caption{Motivating example. The ``sa-rec'' denotes the $(s,a)$-rectangular while the ``d-rec'' denotes the $d$-rectangular adopted by this paper. The ``Proj. sa-rec'' denotes the projected $(s,a)$-rectangular values into the linear space w.r.t. $\phi(s,a)$, which leads to the wrong direction. See Appendix~\ref{app:toy_example} for details.}
    \label{fig:motivation}
\end{figure}

As shown by \citet{wiesemann2013robust},
computing an optimal robust policy
for general ambiguity set is strongly NP-Hard.
For computational tractability, \citet{nilim2005robust, iyengar2005robust} introduce the $(s,a)$-rectangular ambiguity set,
where $\gP_h = \prod_{(s,a)\in \gS\times\gA} \gP_h(s,a)$ implies that the transition probability $P$ taking values independently for each state-action pair. 
The assumption ensures the perturbation of the transition probability for each $(s,a)$-pair is unrelated to others.
However, the $(s,a)$-rectangular assumption may be prohibitively time-consuming when the state space or action space is large, as it requires to solve the robust optimization problem for each $(s,a)$-pair.
More importantly, it may suffer over-conservatism, especially when the transition probabilities enjoy some inherent structure \citep{wiesemann2013robust, goyal2022robust}.

Prior to this work, the only attempts in linear function approximation for the robust RL are \citet{tamar2013scaling, badrinath2021robust}, both sharing the same idea, i.e., first to obtain the robust value for each $(s,a)$-pair and then use a set of linear function basis to approximate the robust values for the whole $\gS\times\gA$.
We use a motivating example with the inherent structure to compare our proposed algorithm with theirs to point out that directly projecting the robust values into the linear space might result in a poor approximation.
We consider a continuous bandit case, which corresponds to the case of offline RL with $H=1$ and $S=1$. 
The action set is $[0, 1]$. 
When taking the action $a=0$, we receive a reward $r_0$ drawing from $\mathcal{N}(1, 1)$. 
If the action $a=1$ are chosen, the reward $r_1$ follows $\mathcal{N}(0, 0.5^2)$. 
For $a \in (0,1)$, its reward distribution $r_a$ is the mixture of $r_0$ and $r_1$, which means the reward follows $\mathcal{N}(1, 1)$ w.p. $1-a$ and follows $\mathcal{N}(0, 0.5^2)$ w.p. $a$.

Since the problem enjoys a linear structure, using the linear function approximation to maintain the low-dimension representation is desirable.
However, as shown in Figure~\ref{fig:motivation}, 
the projected robust values using \citet{tamar2013scaling, badrinath2021robust}'s methods are irrational due to the nonlinear nature of the $(s,a)$-uncertainty set. 
In particular, the projected algorithm behaves even more pessimistically than the $(s,a)$-rectangular for the actions close to $0$ or $1$. 
More importantly, it fails to preserve the order relationship between the robust values of actions $0$ and $1$, i.e., the robust value of action $0$ is higher than $1$ in the $(s,a)$-rectangular but becomes lower after projection, which will potentially lead to wrong decisions.
Instead, our proposed algorithm based on the $d$-rectangular ambiguity set (see Section~\ref{sec:linear-func} for definition),  
recovers the robust values for action $0$ and $1$ while avoiding the over-conservatism of $a \in (0,1)$.



\section{Linear Function Approximation}\label{sec:linear-func}In this section, we assume the MDP enjoys the soft state-aggregation structure and introduce the so-called $d$-rectangular ambiguity set.
We then propose our first algorithm, Distributional Robust Value Iteration with Linear function approximation (DRVI-L) with corresponding non-asymptotic suboptimality results. 
\begin{assumption}[State-Aggregation MDP]
    \label{as:factor-mdp}
    The true transition model is soft state-aggregation w.r.t. $\phi$ and the reward function admit linear representation w.r.t. $\phi$ (Definition~\ref{def:linear}), i.e., $r_h\in \operatorname{Span}(\phi)$.
\end{assumption}
\subsection{Ambiguity Set Structure}
As shown in the motivating example, the $(s,a)$-rectangularity assumption can potentially yield over-conservative policy, especially when the problem enjoys some inherent structure.
In this paper, we consider the problem with a linear structure, i.e., the transition model admits a soft stage aggregation (Definition~\ref{def:linear}), and the reward function also has a linear representation.
Thus building the ambiguity set based on the specific MDP's linear structure might be more natural and less conservative to derive an efficient solution.

To tackle the above challenges, we assume each factor lies in an ambiguity set, which is formally stated in the Assumption~\ref{as:d-rec}.
\begin{assumption}[$d$-rectangular]
\label{as:d-rec}
For each $h\in [H]$, we assume the ambiguity set $\gP_h$ with radius $\rho$ admits the following structure for some probability distance $D:\R^{d}\times \R^{d}\rightarrow \R_{\ge 0}$,
\small{
\begin{align*}
    \gP_h(\rho) =  \left\{ \left(\sum_{i=1}^d \phi_i(s,a)\psi^{\prime}_{h,i}(\spr) \right)_{sa\spr}: \forall D(\psi_{h,i}^{\prime}, \psi_{h,i})\le \rho \right \}.
\end{align*}
}
\end{assumption}

Assumption~\ref{as:d-rec} implies that the factors are unrelated and thus each factor $\psi_{h,i}$ can be chosen arbitrarily within the set $\gP(\psi_{h,i};\rho)\coloneqq\{\psi_{h,i}^{\prime} : D(\psi_{h,i}^{\prime}, \psi_{h,i})\le \rho\}$ without affecting $\psi_{h,j}$ for $j\neq i$.
Our offline DRRL problem with $d$-rectangular is formulated as 
\begin{equation}
\label{eq:dist_robust_offline_RL}
    R(\pi;\gP) = \inf_{P\in \gP(\rho)=\prod_{h=1}^{H}\gP_h(\psi_{h,i}; \rho)} R(\pi,P).
\end{equation}
To provide a concrete algorithm design and corresponding suboptimality analysis, we choose $D$ as the KL divergence.
The corresponding ambiguity for horizon $h$ and the $i$-th factor is denoted as $\gP^{\operatorname{KL}}(\psi_{h,i}; \rho)$ and the ambiguity set for horizon $h$ is denoted as $\gP^{\operatorname{KL}}(\psi_{h}; \rho)\coloneqq \prod_{i\in[d]}\gP^{\operatorname{KL}}(\psi_{h,i}; \rho)$.
\citet{hu2013kullback} proves a strong duality lemma to ensure the computational tractability of the distributionally robust optimization with KL divergence in Equation~\ref{eq:dist_robust_offline_RL}.
\begin{lemma}[\citet{hu2013kullback}]
    \label{lem:huKL}
    Suppose $X\sim P$ has finite moment generating function in the neighborhood of zero. Then ,
    {\small
    \begin{align}
        \inf_{P^{\prime}: D_{\rm KL}(P^{\prime} \parallel P)\le \rho}  \E_{P^\prime}[X] &=  \sup_{\beta\ge 0}\{-\beta \log(Z)-\beta \cdot \rho\}\notag \\
        & = \sup_{\beta\ge 0} \sigma(Z, \beta)\label{eq:dual_operator_KL},
    \end{align}
    }
    where $Z:=\E_{P}[e^{-X/\beta}]$.
\end{lemma}
When $\rho\to 0$, the LHS degrades to the non-robust view, i.e., $\E_{P}[X]$, and the optimum $\beta^{*} = \arg \sup_{\beta\ge 0} \sigma(Z,\beta)$ goes to infinity. 


Based on Lemma~\ref{lem:huKL}, we can derive the following DR Bellman operator:
\begin{equation}
    \label{eq:Bellman}
    \begin{aligned}
    & (\sB_h V)(s,a)= r(s,a) + \inf_{P_h\in \gP^{\operatorname{KL}}( \psi_h;\rho)}\E_{s^{\prime}\sim P_h(\cdot\lvert s,a)}[V(s^{\prime})]\\
    & = \phi(s,a)^{\top} (\theta_h + w_h) = \phi(s,a)^{\top} \nu_h,
    \end{aligned}
\end{equation}
where $w_{h,i} = \sup_{\beta \ge 0}\sigma(\mu_{h,i},\beta)$ and \\
$\mu_{h,i} \coloneqq \mathbb{E}_{\psi_{h,i}}[e^{-V(s')/\beta}]$.

As shown above, the DR Bellman operator using the $d$-rectangular ambiguity set can maintain $\phi$ representation.
We formalize it in Lemma~\ref{lemma:span} and based upon it, we design our DR variant of the value iteration algorithm.
\begin{lemma}
    \label{lemma:span}
    For any policy $\pi$ and any epoch $h\in [H]$, the distributional robust Q-function is linear w.r.t. $\phi$. 
    Moreover, $d(\sB_h \gF, \gF) = 0$,
    where $d(\sB_h\gF, \gF)=\sup_{g\in \gF}\inf_{f\in \gF}\lVert f - \sB_hg\rVert$ is the Bellman error \citep{munos2008finite}.
\end{lemma}

\subsection{Distributionallly Robust Value Iteration}
We are in the position to introduce the offline DRRL setting.
The key challenge in the offline RL is that we restrict the computation of the policy with only access to some logged dataset instead of knowing the exact transition probability.
Due to the lack of continuing interaction with the environment,
the performance of the offline RL algorithm suffers from the insufficient coverage of the offline dataset.
As a start, we first consider the uniformly ``well-explored'',
which is adopted widely in many offline RL works~ \cite{jin2021pessimism,duan2019state,xie2021bellman}.

\begin{assumption}[Uniformly Well-explored Dataset]
    \label{as:sufficient-coverage}
    Suppose $\mathcal{D}$ consists of $N$ trajectories $\{(s_{h}^{\tau}, a_h^{\tau}, r_{h}^{\tau})\}_{\tau,h=1}^{N,H}$ independently and identically induced by a fixed behavior policy $\overline{\pi}$ in the linear MDP. Meanwhile, suppose there exists an absolute constant $\underline{c}>0$ such that at each step $h\in[H]$,
    \begin{align*}
      \lambda_{\min}(\Sigma_h)\ge \underline{c}, \quad \text{ where } \Sigma_h = \E_{\overline{\pi}}[\phi(s_h,a_h)\phi(s_h,a_h)^{\top}].
    \end{align*}
\end{assumption}
Such an assumption requires the behavior policy to explore each feature dimension well, which might need to explore some state-action pairs that the optimal policy has seldom visited.

In principle, we construct the empirical version of the Bellman operator (see Equation~\ref{eq:Bellman}) to approximate the true one, in particular, to approximate $\mu_{h,i}$.
However, we cannot directly observe the samples from the distribution $\psi_{h,i}$ and thus fail to approximate it directly and estimate $\mu_{h,i}$.
Instead, notice that 
\begin{small}
\begin{equation*}    
\mathbb{E}_{P_{s,a}}[e^{-V(s')/\beta}] = \int_{s'}e^{-V(s')/\beta}P(s'\lvert s,a)ds' =\phi(s,a)^{\top}\mu_h,
\end{equation*}
\end{small}
and samples from $P_{s,a}$ can be obtained, which motivates us to approximate $\mu_h$ by linear regression.
We define the empirical mean squared errors as
\begin{align}
    \label{eq:msbe_reward}
    \gE_{h,1}(\theta) &= \sum_{\tau = 1}^K (r_h^{\tau}  - \phi(s_h^{\tau}, a_h^{\tau})^{\top}\theta)^2, \\
    \label{eq:msbe_ambiguity}
    \gE_{h,2}(\mu) &= \sum_{\tau=1}^{K} (e^{-V(s_{h+1}^{\tau})/\beta} - \phi(s_h^{\tau}, a_h^{\tau})^{\top}\mu)^2,
\end{align}
at each step $h\in [H]$.
Correspondingly, we have 
\begin{align*}
    (\widehat{\sB}_h V)(s,a) &= \phi(s,a)^{\top} \widehat{\theta}_h + \phi(s,a)^{\top} \widehat{w}_h, \\
    \text{where} \quad \widehat{\theta}_h &= \arg\min_{\theta\in \sR^d} \gE_{h,1}(\theta) + \lambda\cdot \lVert \theta\rVert^2,\\
    \widehat{\mu}_h &= \arg\min_{\mu\in \sR^d} \gE_{h,2}(\mu) + \lambda\cdot \lVert \mu\rVert^2,\\
    \widehat{w}_{h,i} &= \sup_{\beta\geq 0}\sigma(\widehat{\mu}_{h,i}, \beta).
\end{align*}
Here $\lambda>0$ is the regularization parameter.



Note that ridge penalization while ensuring numerical safe, inducing the solution $\widehat{\mu}_h$ to get close to zero. 
A close to zero $\widehat{\mu}_{h,i}$ would render $\sigma(\widehat{\mu}_{h,i}, \beta)$ to approach infinity and destroy our estimation.
Thus, we introduce a novel value shift technique in the $\widehat{w}_h$ by defining 
\begin{align}
\label{eq:dual_operator_KL_shift}
    \tilde{\sigma}(Z, \beta)  = -\beta \log(Z+1)-\beta \cdot \rho,
\end{align}
to ensure $\log(Z+1)$ still remain valid even $Z$ approach zero.
Correspondingly, we adopt the shifted variant of regression 
objective (see Equation~\ref{eq:msbe_ambiguity}), defined as
\begin{align}
    \tilde{\gE}_{h,2}(\mu) &= \sum_{\tau=1}^{K} ((e^{-V(s_{h+1}^{\tau})/\beta} - 1) - \phi(s_h^{\tau}, a_h^{\tau})^{\top}\mu)^2, \notag \\
    \widehat{\mu}_h &= \arg\min_{\mu\in \sR^d} \tilde{\gE}_{h,2}(\mu) + \lambda\cdot \lVert \mu\rVert^2, \notag \\
    \widehat{w}_{h,i} &= \sup_{\beta\ge 0}\tilde{\sigma}(\widehat{\mu}_{h,i}, \beta). \label{eq:msbe_ambiguity_shift}
\end{align}
$\widehat{\theta}_h$ and $\widehat{w}_h$ have the closed form 
\begin{small}
\begin{align*}
    &\widehat{\theta}_h = \Lambda_h^{-1} \left(\sum_{\tau=1}^K \phi(s_h^{\tau}, a_h^{\tau})r_h^{\tau} \right), \\
    &\widehat{\mu}_h = \Lambda_h^{-1} \left(\sum_{\tau=1}^{K} \phi(s_h^{\tau}, a_h^{\tau}) (e^{-V(s_{h+1}^{\tau}/\beta)} - 1)\right),\\
    &\text{where}\quad \Lambda_h = \sum_{\tau=1}^K \phi(s_h^{\tau}, a_h^{\tau}) + \lambda \cdot I.
\end{align*}
\end{small}

Our value shifting technique can ensure $\widehat{w}_h$ maintains a valid value no matter the quality of the estimator, which is the key to achieving the desired suboptimality that can nearly recover to that of the non-robust setting.
We summarize our algorithm as \textbf{D}istributional \textbf{R}obust \textbf{V}alue \textbf{I}teration with \textbf{L}inear function approximation (DRVI-L) in Algorithm~\ref{alg:drvi}. 




Before we present the suboptimality analysis for our algorithm~\ref{alg:drvi}, we impose the following Assumption~\ref{as:lower-bound-beta} to assume a common known lower bound for the optimum of the KL optimization problem in Lemma~\ref{lem:huKL}.
Such an assumption is also needed in the tabular case \citep{zhou2021finite}.

\begin{assumption}
    \label{as:lower-bound-beta}
    For each $h\in [H]$ and each $i\in [d]$, we denote 
    \begin{align*}
        \beta_{h,i}^{*} = &\operatorname{\arg\sup}_{\beta_{h,i}\ge 0}\sigma(\mu_{h,i},\beta_{h,i}).
    \end{align*}
    We assume there is a common lower bound for $\{\beta_{h,i}^{*}\}_{h,i=1}^{H,d}$, i.e., we assume there exists a known $\underline{\beta}$ s.t. $0<\underline{\beta}\le  \min_{h\in [H], i\in [d]}\beta_{h,i}^{*}$.
\end{assumption}
By Proposition 2 in \cite{hu2013kullback}, $\beta_{h,i}^{*} = 0$ when the worst case happens with sufficient large probability w.r.t. $\rho$, i.e., the larger the $\rho$, the easier we have $\beta_{h,i}^{*} = 0$.
In practice, we usually use a small $\rho$ to adapt to the problem without incurring over-conservatism, which utilize the advantage of DRO compared with other robustness notions \cite{ben1998robust,ben2000robust, duchi2021learning}.

\begin{algorithm}[ht]
    \caption{DRVI-L}
    \label{alg:drvi}
   \begin{algorithmic}[1]
      \STATE {\bfseries Input:} $\underline{\beta}$, $ \mathcal{D}=\left\{\left(s_{h}^{\tau}, a_{h}^{\tau}, r_{h}^{\tau}\right)\right\}_{\tau, h=1}^{N, H}$.
      \STATE {\bfseries Init:} $\widehat{V}_{H} = 0$.
      \FOR {step $h=H$ {\bfseries to} $1$}
      \STATE $\Lambda_{h}
      \leftarrow \sum_{\tau=1}^{N} \phi\left(s_{h}^{\tau}, a_{h}^{\tau}\right) \phi\left(s_{h}^{\tau}, a_{h}^{\tau}\right)^{\top}+\lambda I$
      \STATE $\widehat{\theta}_h \gets \Lambda_h^{-1} \left[\sum_{\tau=1}^{N} \phi(s_h^\tau, a_h^\tau) r_h^\tau \right]$
      \IF {$h=H$}
      \STATE $\widehat{w}_H \gets 0$
      \ELSE 
    \STATE Update $\widehat{w}_{h,i} $ with Equation~\ref{eq:msbe_ambiguity_shift}.
       \ENDIF
       \STATE $\widehat{\nu}_h = \min(\widehat{\theta}_h + \widehat{w}_h, H-h+1)_+$
      \STATE $\widehat{Q}_h(\cdot,\cdot) \gets \phi(\cdot,\cdot)^\top \widehat{\nu}_h$
     \STATE $\widehat{\pi}_{h}(\cdot \mid \cdot) \leftarrow \arg \max _{\pi_{h}}\langle\widehat{Q}_{h}(\cdot, \cdot), \pi_{h}(\cdot \mid \cdot)\rangle_{\mathcal{A}}$
     \STATE $\widehat{V}_{h}(\cdot) \leftarrow\langle\widehat{Q}_{h}(\cdot, \cdot), \widehat{\pi}_{h}(\cdot \mid \cdot)\rangle_{\mathcal{A}}$
      \ENDFOR
   \end{algorithmic}
\end{algorithm}

\begin{theorem}
    \label{thm:sufficient-coverage}
    We set $\lambda =1$ in Algorithm~\ref{alg:drvi}. 
    Under the Assumption~\ref{as:factor-mdp}, Assumption~\ref{as:sufficient-coverage} and Assumption~\ref{as:lower-bound-beta}, when $N\ge 40/\underline{c}\cdot \log(4dH/\delta)$, we have the following holds with probability at least $1-\delta$,
    \begin{align*}
        \operatorname{SubOpt}(\widehat{\pi};\gP)&\le c_1 \underline{\beta}(e^{H/\underline{\beta}}-1) d^{1/2}\zeta_1^{1/2}H/N^{1/2} \\
        &+ c_2 \underline{\beta}^{1/2}(e^{H/\underline{\beta}}-1)\zeta_2^{1/2}H^{3/2}/N^{1/2}.
    \end{align*}
    Here $\zeta_1 = \zetaI$, $\zeta_2=\zetaII$ and $c_1$ and $c_2$ are some absolute constants that only depend on $\underline{c}$.
\end{theorem}

Notice that the suboptimality of algorithm~\ref{alg:drvi} mainly depends on the dimension $d$ instead of the size of the state-action space.
Compared with tabular cases, e.g., \citep{zhou2021finite, yang2021towards}, which mainly bounds the finite sample error for each $(s,a)$ pair separately, Theorem~\ref{thm:sufficient-coverage} is derived by exploiting the linear structure shared by various $(s,a)$-pairs, building a novel $\epsilon$-net to control the finite sample error for whole linear function family space, and utilizing the power of our value shift algorithmic ingredient.
The DRO ingredient requires to design a specific $\epsilon$-net for the function family that is different from the non-robust linear function approximation.

In particular, 
$\tilde{O}(d^{1/2}H^{3/2}\sqrt{\underline{\beta}}e^{H/\underline{\beta}}/N^{1/2})$ when $\underline{\beta}$ is relatively small, i.e., when the algorithm tends to learn a pessimistic view.
When $\underline{\beta}\rightarrow \infty$, i.e., the algorithm is learning a nearly non-robust view, our bound reduces to $\tilde{O}(d^{1/2}H^2/N^{1/2})$, which recovers to the same dependence on $H$  compared with the non-robust PEVI algorithm in \citet{jin2021pessimism} and achieve the optimal dependence on $N$ and $d$ in \citet{yin2022nearoptimal}.
The gap in the $H$ is because we adopt a relatively simple algorithmic design framework to outline the first step in DRRL linear function approximation.
Adopting the advance technique of \citet{yin2022nearoptimal} may solve the gap and we leave it as a further direction.




\section{Extensions}\label{sec:extension}

\subsection{Beyond Uniformly Well-explored Dataset}
In real applications, the data coverage may not be ideal as Assumption~\ref{as:sufficient-coverage}, which requires the behavior policy to explore all the feature dimensions with a sufficiently high exploration rate. 
Instead, we only require the behavior policy to have sufficient coverage of the features that the optimal policy will visit.
Thus we design the pessimistic variant of our Algorithm~\ref{alg:drvi}, called the \textbf{P}essimistic \textbf{D}istributionally \textbf{R}obust \textbf{V}alue \textbf{I}teration with \textbf{L}inear function approximation (PDRVI-L), by leveraging the idea of~\citet{jin2021pessimism}. 
Under a weaker data coverage condition, sample efficiency can be obtained as long as the trajectory induced by the optimal policy $\pi^{*}$ is sufficiently covered by the dataset sufficiently well.
We formalize this condition in Assumption~\ref{as:partial-coverage}.
\begin{assumption}[Sufficient Coverage of the Optimal Policy]
    \label{as:partial-coverage}
    Suppose there exists an absolute constant $c^{\dagger}>0$ such that 
    \begin{small}
    \begin{align*}
        \Lambda_h\ge I + c^{\dagger}\cdot N\cdot d\cdot \E_{\pi^{*}}[(\phi_i(s_h,a_h)\mathds{1}_i)(\phi_i(s_h,a_h)\mathds{1}_i)^{\top}\lvert s_1=s], 
    \end{align*}
    \end{small}
    $\forall s\in \gS, h\in[H], i\in[d]$, holds for probability at least $1-\delta$.
\end{assumption}
Compared with sufficient coverage condition in \cite{jin2021pessimism}, our Assumption~\ref{as:partial-coverage} requires the collected samples $\Lambda_h$ covers uniformly well for different dimension $i\in[d]$. 
Such a requirement comes from the ambiguity set constructed from the feature space.
We summarize our algorithmic design in Algorithm~\ref{alg:pdrvi}, which is deferred to Appendix~\ref{appendix:pdrvi}.
Compared with Algorithm~\ref{alg:drvi}, we subtract a pessimistic term $\gamma_h \sum_{i=1}^d \lVert \phi_i(s,a)\mathds{1}_i\rVert_{\Lambda_h^{-1}}$ from the estimated $Q$-value.
This pessimistic term discourages our algorithm from choosing the action with less confidence.
Compared with \cite{jin2021pessimism} which use 
$\gamma_h\lVert \phi(s,a)\rVert_{\Lambda_h^{-1}}$ as the pessimistic term in non-robust setting, ours is a larger penalization and adapt to the distributionally robust nature.
Under the partial coverage condition for our dataset, our Algorithm~\ref{alg:pdrvi} enjoys sample efficiency as concluded in the following theorem.

\begin{theorem}
    \label{thm:partial-coverage}
    In Algorithm~\ref{alg:pdrvi} we set $\lambda=1$ and 
    \begin{align*}
        \gamma_h &= \gammah,
    \end{align*}
    where $\zeta_2$ is the same as in Theorem~\ref{thm:sufficient-coverage} and $\zeta_3 = \zetaIII$
    for some absolute constant $c_1$ and $c_2$ that are only dependent on $c^{\dagger}$. 
    Then under the Assumption~\ref{as:factor-mdp}, \ref{as:lower-bound-beta} and \ref{as:partial-coverage},
    our algorithm~\ref{alg:pdrvi} has the following guarantee with probability at least $1-\delta$,
    \begin{align*}
        \operatorname{SubOpt}(\widehat{\pi};\gP)\le & c_1 \underline{\beta}(e^{H/\underline{\beta}}-1)  d^{3/2}H\zeta_3^{1/2}/N^{1/2} \\+ &c_2\underline{\beta}^{1/2}(e^{H/\underline{\beta}}-1)d^{1/2}H^{3/2}\zeta_2^{1/2}/N^{1/2}.
    \end{align*}
\end{theorem}
This bound incurs an extra $d$ compared with Theorem~\ref{thm:model-misspecific-sufficient} as a price to work for a weaker data coverage condition.
In specific, the suboptimality for the Algorithm~\ref{alg:pdrvi} is $\tilde{O}(d^{3/2}H^{3/2}\sqrt{\underline{\beta}}e^{H/\underline{\beta}}/N^{1/2})$ when $\underline{\beta}$ is relatively small.
When $\underline{\beta}\rightarrow \infty$, i.e., the algorithm is learning a nearly non-robust view, the suboptimality reduces to $\tilde{O}(d^{3/2}H^2/N^{1/2})$, which recovers to the same dependence of $d$, $H$ and $N$ as~\citet{jin2021pessimism}.
Recently,~\citet{yin2022nearoptimal} improves the suboptimality bound to $\tilde{O}(d^{1/2}H^{3/2}/N^{1/2})$ with a complicate algorithm design.
Instead, our paper is the first attempt to design linear function approximation to solve the distributionally robust offline RL problem. Thus we leave the improvement to the optimal rate as a further direction.


\begin{figure*}[htbp]
\centering
\hfill
\subfigure[Average Return]{
\label{fig.sub.1}
\includegraphics[width=0.31\textwidth]{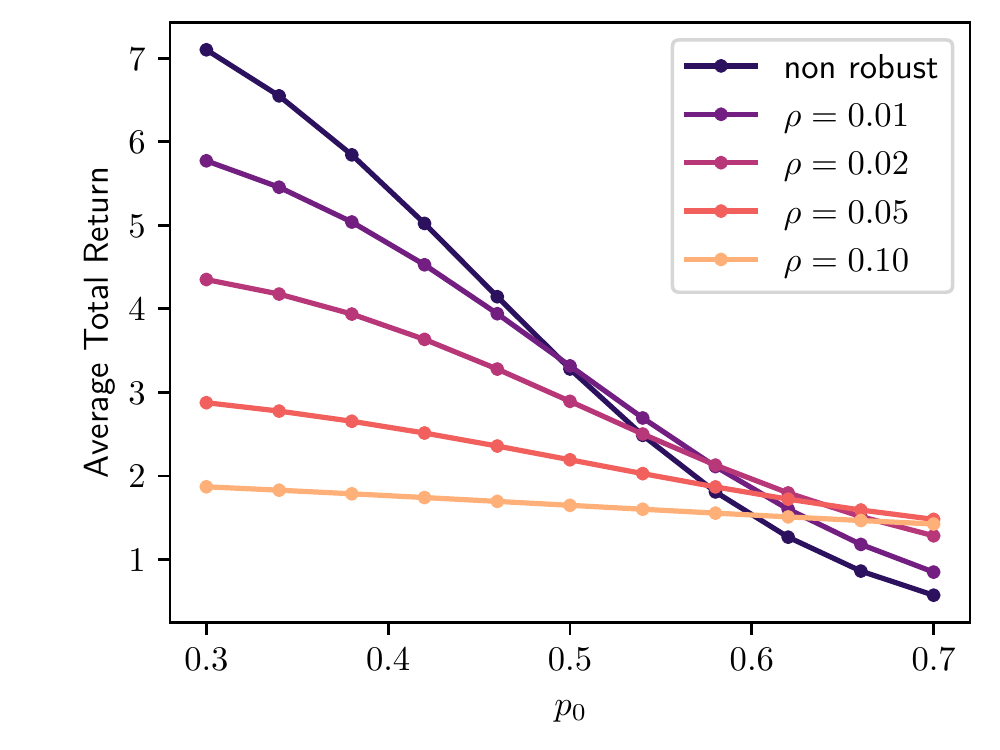}} \hfill
\subfigure[$\| \widehat{V}_1 - V^*_1\|$]{
\label{fig.sub.2}
\includegraphics[width=0.31\textwidth]{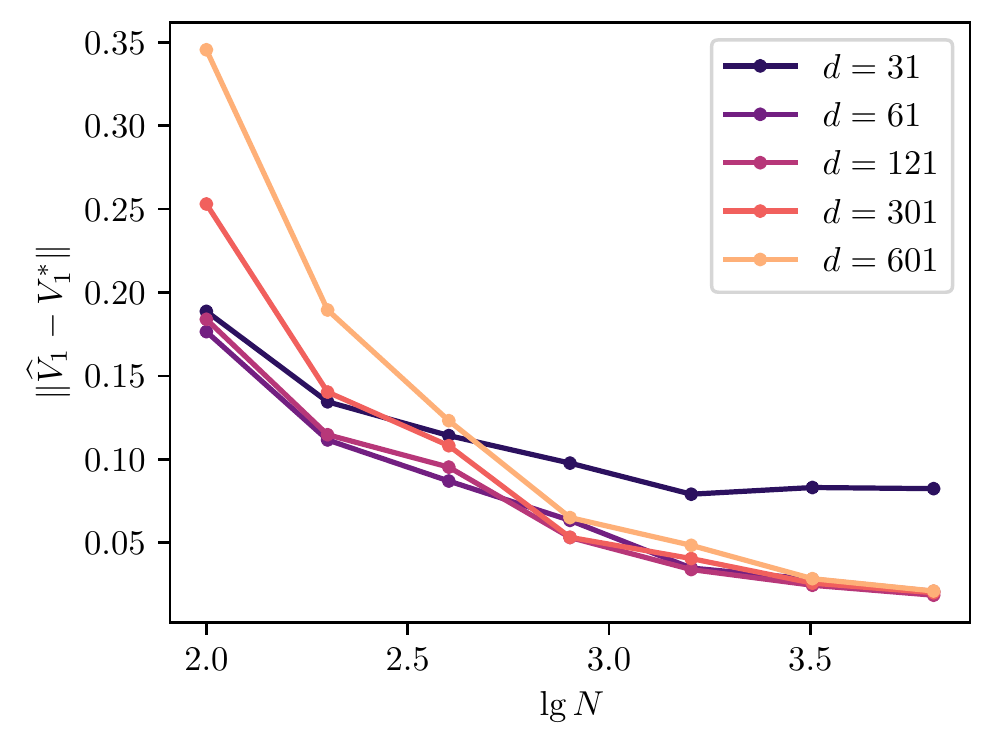}} \hfill
\subfigure[Execution time]{
\label{fig.sub.3}
\includegraphics[width=0.31\textwidth]{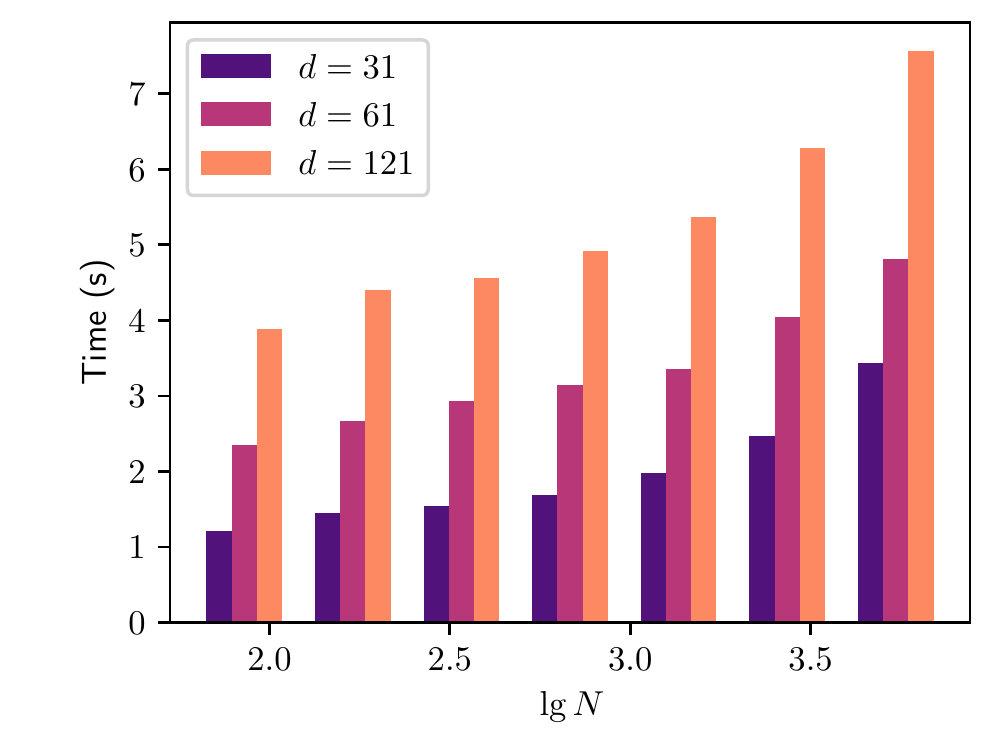}}
\hfill
\caption{Results in American Option Experiment. (a) Average total return of different KL radius $\rho$ in the perturbed environments ($N = 1000$, $d=61$). (b) Estimation error with different linear function dimension $d$'s and the sizes of dataset $N$'s ($\rho=0.01$).  (c) Execution time for different $d$'s.}
\label{fig.lable}
\end{figure*}

\subsection{Model Misspecification}
The state aggregation assumption may not be realistic when applied to the real-world dataset. 
In this subsection, we relax the assumption of the soft state-aggregation MDP to allow the true transition kernel is nearly a state-aggregation transition.
\begin{assumption}[Model Misspecification in Transition Model]
    \label{as:misspecification}
    We assume that for all $h\in[H]$, there exists $\tilde{P}_h\in \operatorname{Span}(\phi)$ and $\xi\ge 0$ such that each $(s,a)$, the true transition kernel $P_h(\cdot\lvert s,a)$ satisfies 
    $\lVert P_h(\cdot \lvert s,a) - \tilde{P}_h(\cdot \lvert s,a) \rVert_1\le \xi$.
    For the reward functions, without loss of generality we still assume that $r_h\in \operatorname{Span}(\phi)$ for all $h\in[H]$.
\end{assumption}

\begin{theorem}[Model Misspecification]
    \label{thm:model-misspecific-sufficient}
We set $\lambda =1$ in Algorithm~\ref{alg:drvi}. 
    Under the Assumption~\ref{as:sufficient-coverage}, \ref{as:lower-bound-beta} and \ref{as:misspecification}, when $N\ge 40/\underline{c}\cdot \log(4dH/\delta)$, we have the following holds with probability at least $1-\delta$,
    \begin{align*}
        &\operatorname{SubOpt}(\widehat{\pi};\gP)\le c_1 \underline{\beta}(e^{H/\underline{\beta}}-1)(\xi d^{1/2} + d^{1/2}\zeta_1^{1/2})H/N^{1/2} \\
        & \quad + c_2 \underline{\beta}^{1/2}(e^{H/\underline{\beta}}-1)H^{3/2}\zeta_2^{1/2}/N^{1/2} + H(H-1)\xi/2.
    \end{align*}
Here $\zeta_1$ and $\zeta_2$ are the same in Theorem~\ref{thm:sufficient-coverage} and $c_1$ and $c_2$ are some absolute constants that only depend on $\underline{c}$.
\end{theorem}

\begin{theorem}[Model Misspecification with Sufficient Coverage]
    \label{thm:model-misspecific-insufficient}
    In Algorithm~\ref{alg:pdrvi} we set $\lambda=1$ and 
    \begin{align*}
        \gamma_h &= \gammah,
    \end{align*}
    where $\zeta_2$ and $\zeta_3$ are the same as in Theorem~\ref{thm:partial-coverage} and $c_1, c_2\ge 1$ are some absolute constants that only involve $c^{\dagger}$.
    Then based on Assumptions~\ref{as:lower-bound-beta}, \ref{as:partial-coverage} and \ref{as:misspecification}, our Algorithm~\ref{alg:pdrvi} has the following guarantee with probability at least $1-\delta$, 
    \begin{align*}
        & \operatorname{SubOpt}(\widehat{\pi};\gP)\le c_1 \underline{\beta}(e^{H/\underline{\beta}}-1)(\xi d +  d^{3/2}\zeta_3^{1/2})H/N^{1/2} \\
        &+ c_2\underline{\beta}^{1/2}(e^{H/\underline{\beta}}-1)d^{1/2}H^{3/2}\zeta_2^{1/2}/N^{1/2} + H(H-1)\xi/2.
    \end{align*}
\end{theorem}

Theorem~\ref{thm:model-misspecific-sufficient} 
implies that when the soft-state aggregation model is inexact up to $\xi$ total variation, there is an approximation gap in the policy's performance $O(\xi \cdot (\underline{\beta}(e^{H/\underline{\beta}} - 1)d^{1/2} + H^2))$ and $O(\xi \cdot (\underline{\beta}(e^{H/\underline{\beta}} - 1)d + H^2))$ for our DRVI-L and PDRVI-L algorithms respectively. 
The level of degradation depends on the total-variation divergence between the true transition distribution and the robustness level we aim to achieve.

\section{Experiment}\label{sec:experiment}
We conduct the experiments to answer: (a) with linear function approximation, does distributionally robust value iteration perform better than the standard value iteration in the perturbed environments? (b) how does the linear dimension $d$ affect performance? 
We evaluate our algorithm in the American put option environment~\citep{tamar2013scaling,zhou2021finite}. 
The experiment setup is classical and is deferred in the Appendix~\ref{appendix:experiment}.

We first compare distributionally robust and standard value iteration with the same linear function approximation. For both algorithms, we use $d=31$ and collect a dataset with 1000 trajectories in the environment with $p_0=0.5$. After training with different uncertainty ball radius $\rho$'s, we evaluate their average performance in the perturbed environment with different $p_0$'s. The results are depicted in Figure~\ref{fig.sub.1}. As expected, the robust agent performs better than the non-robust one better in the perturbed environment with $p>0.55$. We also notice the policy with $\rho=0.01$ has a similar performance with the non-robust one in the nominal environment ($p_0=0.5$) and a much more robust performance across the perturbed environments, showing the superiority of our method.

Next, we investigate the effect of the linear function dimension. We fix the number of trajectories and compare $\|\widehat{V}_1 - V^*_1 \|$ with different $d$'s. We repeat each experiment 20 times and show the average results in Figure~\ref{fig.sub.2}. As our theoretical analysis points out, the smaller $d$ leads to a lower estimation error and higher approximation error with the same data size. The misspecification of the linear transition model will bring an intrinsic bias to the value estimation. Yet, the proper bias brings lower error with limited data, which is essential for offline learning.

We show the computational efficiency of the proposed method, which is measured by the average execution time (see Figure~\ref{fig.sub.3}). The result shows that the average execution time linearly depends on the $d$, which indicates that the computing bottleneck of the algorithm is solving the dual problems (line 9 in Algorithm~\ref{alg:drvi}). All the experiments are finished on a server with an AMD EPYC 7702 64-Core Processor CPU.

\begin{figure}[htbp]
\centering
\includegraphics[width=0.31\textwidth]{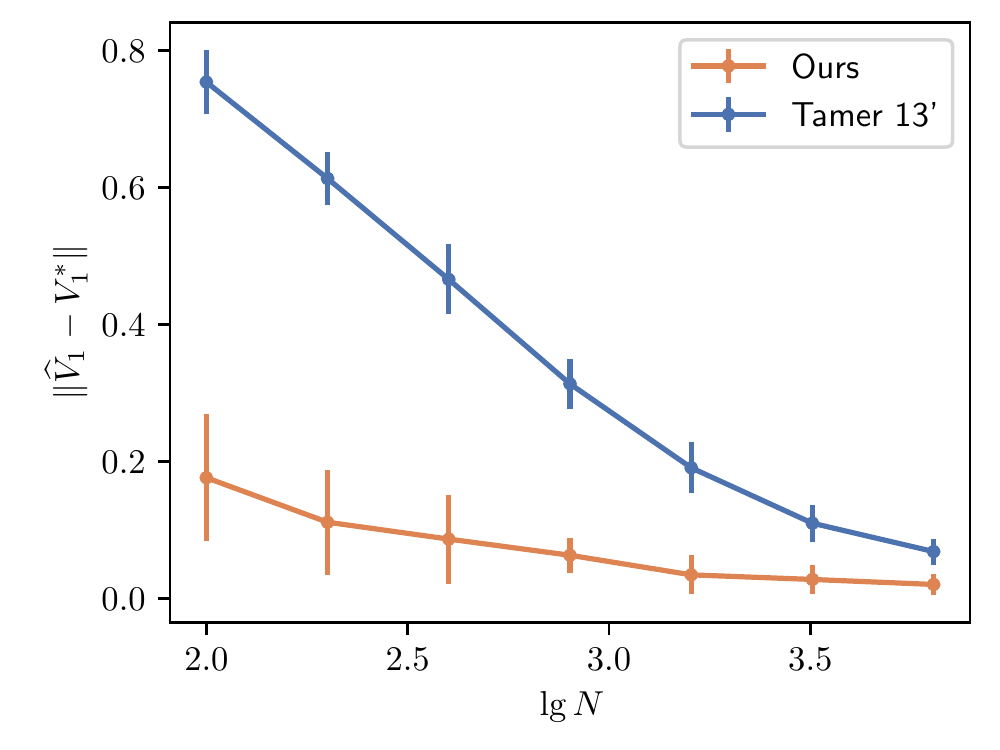}
\caption{Compare the estimation error with RPVI.}
\label{fig.ablation}
\end{figure}

Finally, we compare our method with the robust projected VI (RPVI)~\cite{tamar2013scaling}, in which the robust values are first calculated for each $(s,a)$-pair and then projected into the linear function space. Under the same function class with $d=61$, we compare their value difference with the optimal value in Figure~\ref{fig.ablation}, where our algorithm has more accurate estimations with the varying $N$. In addition, since RPVI solves the robust problem for each $(s,a)$, its execution time linearly depends on the sample size $N$. For instance, with $d=61$ and $N=1000$, RPVI needs over $1000s$, while our DRVI-L can finish less than $10s$.
The details of reproduction are shown in Appendix~\ref{app:RPVI}

\bibliography{reference}
\bibliographystyle{icml2023}

\newpage
\appendix
\onecolumn

\section{Details of the motivating example} \label{app:toy_example}
For convenience, given a r.v. $X$, we denote its distributional robust value (refer Lemma~\ref{lem:huKL}) as
$g(X, \rho)=\sup_{\beta\ge 0}\{-\beta \log(\E_{X\sim P}[e^{X/\beta}]) - \beta \cdot \rho\}.$ For any action $a$, the corresponding reward distribution is
\begin{equation}
r_a \sim \begin{cases}
\mathcal{N}(1, 1)  & {\rm w.p.} \quad 1-a \\ 
\mathcal{N}(0, 0.5) &  {\rm w.p.} \quad a
\end{cases}
\end{equation}
Based on the definition of $(s,a)$-rectangular, the robust action-value function $Q_{\rm sa}(a)=g(r_a, 1)$.
The projected value function is approximated with a linear function $Q_{\rm proj}(a)=[1-a; a]^\top w $, where $w = \argmin_{u} \E_{a \sim U[0,1]} (V_{\rm sa}(a) - [1-a; a]^T u)^2$.
The $d$-rectangular robust action-value function is $Q_{\rm d}(a)= (1-a)g(r_0, 1)+ag(r_1, 1)$.

\section{Algorithm Design for the PDRVI-L}
\label{appendix:pdrvi}

\begin{algorithm}[ht]
    \caption{PDRVI-L}
    \label{alg:pdrvi}
   \begin{algorithmic}[1]
      \STATE {\bfseries Input:} $\underline{\beta}$, $ \mathcal{D}=\left\{\left(s_{h}^{\tau}, a_{h}^{\tau}, r_{h}^{\tau}\right)\right\}_{\tau, h=1}^{N, H}$.
      \STATE {\bfseries Init:} $\widehat{V}_{H} = 0$.
      \FOR {step $h=H$ {\bfseries to} $1$}
      \STATE $\Lambda_{h}
      \leftarrow \sum_{\tau=1}^{N} \phi\left(s_{h}^{\tau}, a_{h}^{\tau}\right) \phi\left(s_{h}^{\tau}, a_{h}^{\tau}\right)^{\top}+\lambda I$
      \STATE $\widehat{\theta}_h \gets \Lambda_h^{-1} \left[\sum_{\tau=1}^{N} \phi(s_h^\tau, a_h^\tau) r_h^\tau \right]$
      \IF {$h=H$}
      \STATE $\widehat{w}_H \gets 0$
      \ELSE 
    \STATE Update $\widehat{w}_{h,i} $ with Equation~\ref{eq:msbe_ambiguity_shift}.
       \ENDIF
       \STATE $\widehat{\nu}_h = \min(\widehat{\theta}_h + \widehat{w}_h, H-h+1)_+$
      \STATE $\widehat{Q}_h(\cdot,\cdot) \gets \phi(\cdot,\cdot)^\top \widehat{\nu}_h -  \gamma_h \sum_{i=1}^d \lVert \phi_i(s,a)\mathds{1}_i\rVert_{\Lambda_h^{-1}}$
     \STATE $\widehat{\pi}_{h}(\cdot \mid \cdot) \leftarrow \arg \max _{\pi_{h}}\langle\widehat{Q}_{h}(\cdot, \cdot), \pi_{h}(\cdot \mid \cdot)\rangle_{\mathcal{A}}$
     \STATE $\widehat{V}_{h}(\cdot) \leftarrow\langle\widehat{Q}_{h}(\cdot, \cdot), \widehat{\pi}_{h}(\cdot \mid \cdot)\rangle_{\mathcal{A}}$
      \ENDFOR
   \end{algorithmic}
\end{algorithm}

\section{Experiment Setup}
\label{appendix:experiment}
We assume that the price follows Bernoulli distribution~\citep{cox1979option},
\begin{equation}
s_{h+1}= \begin{cases}c_{u} s_{h}, & \text { w.p. } p_0 \\ c_{d} s_{h}, & \text { w.p. } 1-p_0\end{cases}
\end{equation}
where the $c_u$ and $c_d$ are the price up and down factors and $p_0$ is the probability that the price goes up. The initial price $s_0$ is uniformly sampled from $[\kappa - \epsilon, \kappa + \epsilon]$, where $\kappa=100$ is the strike price and $\epsilon=5$ in our simulation. The agent can take an action to exercise the option ($a_h=0$) or not exercise ($a_h=1$) at the time step $h$. If exercising the option, the agent receives a reward $\max(0, \kappa-s_h)$ and the state transits into an exit state. Otherwise, the price will fluctuate based on the above model and no reward will be assigned. In our experiments, we set $H=20$, $c_u=1.02$, $c_d=0.98$. We limit the price in $[80, 140]$ and discretize with the precision of 1 decimal place. Thus the state space size $|\mathcal{S}|=602$.

Since $Q(s_h, a=1)=\max(0, \kappa-s_h)$ is known in advance, we do not need to do any approximation for $a_h=1$, and only need to estimate $\widehat{Q}(s_h, a=0)=\phi(s_h)^\top \widehat{w}_h$. The features are chosen as $\phi(s_h) = [\varphi(s_h, s_1), \dots, \varphi(s_h, s_d)]^\top$, where $s_1, \dots, s_d$ are selected anchor states and $\varphi(s_h, s_i), \forall i \in [d]$ is the pairwise similarity measure. In particular, we set $s_1 = s_{\rm min}=80, s_d = s_{\rm max=}140$, and $\Delta = s_{i+1} - s_{i}=(s_{\rm max} - s_{\rm min}) / (d-1), \forall i \in [d-1]$. The similarity measure 
$\varphi(s_h, s_i) = \operatorname{max}(0, 1 - |s_h - s_i|/\Delta), \forall i \in [d]$, which is the partition to the nearest anchor states. Before training the agent, we collect data with a fixed behavior policy for $N$ trajectories. Since taking $a_h=1$ will terminate the episode, it is helpless for learning the transition model. Hence, we use a fixed policy to collect data, which always chooses $a_h=0$.

\section{Proof of Section~\ref{sec:linear-func}}
\label{proof:linear}
\begin{lemma}
    \label{lem:extended-value-difference}
    Let $\pi=\{\pi_h\}_{h=1}^H$ and $\pi^{\prime} = \{\pi_h^{\prime}\}_{h=1}^H$ be any two policies and let $\{\widehat{Q}_h\}_{h=1}^H$ be any estimated Q-functions. For all $h\in[H]$, we define the estimated value function $\widehat{V}_h:\mathcal{S}\rightarrow \mathcal{R}$ by setting $\widehat{V}_h(s) = \langle \widehat{Q}_h(s,\cdot), \pi_h(\cdot\lvert s)\rangle_{\mathcal{A}}$ for all $s\in \mathcal{S}$.  For all $s\in \mathcal{S}$, we have 
    \begin{align*}
        \widehat{V}_1(s) - V_1^{\pi^{\prime}}(s) &= \sum_{h=1}^H \E_{\pi^{\prime}}[\iota_h(s_h,a_h)\lvert s_1 = s] + \sum_{h=1}^H \E_{\pi^{\prime}}[\langle \widehat{Q}_1(s,\cdot), \pi_h(\cdot\lvert s_h) - \pi^{\prime}_h(\cdot\lvert s_h)\rangle \lvert s_1 = s],
    \end{align*}
    where 
    $$\iota_{h}(s,a) = \widehat{Q}_h(s,a) - (r_h(s,a) + \inf_{P_{h+1}\in\sP_{h+1}(\cdot\vert s,a)}\E_{s^{\prime}\sim P_{h+1}}[\widehat{V}_{h+1}(s^{\prime})]).
    $$ 
\end{lemma}

\begin{proof}
    \begin{align*}
        \widehat{V}_h(s) - V_h^{\pi^{\prime}}(s) &= \langle \widehat{Q}_h(s,\cdot), \pi_h(\cdot\lvert s)\rangle - \langle Q_h^{\pi^{\prime}}(s,\cdot), \pi_h^{\prime}(\cdot\lvert s)\rangle\\
        &= \langle \widehat{Q}_h(s,\cdot) - Q_h^{\pi^{\prime}}(s,\cdot), \pi_h^{\prime}(\cdot\lvert s)\rangle + \langle \widehat{Q}_h(s,\cdot), \pi_h(\cdot\lvert s) -\pi_h^{\prime}(\cdot\lvert s)\rangle\\
        &= \mathbb{J}_{h}(\widehat{Q}_1(s,\cdot) - Q_h^{\pi^{\prime}}(s,\cdot)) + \xi_h,
    \end{align*}    
    where $\mathbb{J}_{h}(f) = \langle f(s,\cdot), \pi_h^{\prime}(\cdot\lvert s)\rangle$.
    Moreover, we denote 
    $$\iota_{h}(s,a) = \widehat{Q}_h(s,a) - (r_h(s,a) + \inf_{P_{h+1}\in\sP_{h+1}(\cdot\vert s,a)}\E_{s^{\prime}\sim P_{h+1}}[\widehat{V}_{h+1}(s^{\prime})]),
    $$ 
    thus $$\widehat{Q}_h(s,a) =  r_h(s,a) + \inf_{P_{h+1}\in\sP_{h+1}(\cdot\lvert s,a)}\E_{s^{\prime}\sim P_{h+1}}[\widehat{V}_{h+1}(s^{\prime})] + \iota_{h}$$ 
    and 
    $$Q_h^{\pi^{\prime}}(s,a) = r_h(s,a)+\inf_{P_{h+1}\in\sP_{h+1}(\cdot \lvert s,a)} \E_{s^{\prime}\sim P_{h+1}}[V_{h+1}^{\pi^{\prime}}(s^{\prime})],$$ 
    which implies
    \begin{align*}
        \widehat{Q}_h(s,a) - Q_h^{\pi^{\prime}}(s,a) &= \inf_{P_{h+1}\in\sP_{h+1}(\cdot \lvert s,a)}\E_{s^{\prime}\sim P_{h+1}}(\widehat{V}_{h+1}(s^{\prime})-V_{h+1}^{\pi^{\prime}}(s^{\prime})) + \iota_h\\
        &\coloneqq  \tilde{\sP}_h(\widehat{V}_{h+1}-V_{h+1}^{\pi^{\prime}}) + \iota_h,
    \end{align*}
    where $\tilde{\sP}_h[H(s^{\prime})]\coloneqq \inf_{P_{h+1}\in\sP_{h+1}(\cdot \lvert s,a)}\E_{s^{\prime}\sim P_{h+1}}[H(s^{\prime})]$.
    Thus we have \begin{align*}
        \widehat{V}_h(s) - V_h^{\pi^{\prime}}(s) &= (\mathbb{J}_{h}\tilde{\mathbb{P}}_h)(\widehat{V}_{h+1}-V_{h+1}^{\pi^{\prime}}) + \mathbb{J}_{h}\iota_h + \xi_h\\
        \widehat{V}_1(s) - V_1^{\pi^{\prime}}(s) &= (\prod_{h=1}^{H} \mathbb{J}_{h}\tilde{\mathbb{P}}_h)(\widehat{V}_{H+1}-V_{H+1}^{\pi^{\prime}}) + \sum_{h=1}^H(\prod_{i=1}^{h-1} \mathbb{J}_{i}\tilde{\sP}_j)\iota_j + (\sum_{h=1}^H\prod_{i=1}^{h-1})\xi_i\\
         &= \sum_{h=1}^H(\prod_{i=1}^{h-1} \mathbb{J}_{i}\tilde{\sP}_j)\iota_j + (\sum_{h=1}^H\prod_{i=1}^{h-1})\xi_i\\
         &= \sum_{h=1}^H \E_{\pi^{\prime}}[\iota_h(s_h,a_h)\lvert s_1 = s] + \sum_{h=1}^H \E_{\pi^{\prime}}[\langle \widehat{Q}_1(s,\cdot), \pi_h(\cdot\lvert s_h) - \pi^{\prime}_h(\cdot\lvert s_h)\rangle \lvert s_1 = s], 
    \end{align*}
    where the second equality is because $\widehat{V}_{H+1}=V_{H+1}^{\pi^{\prime}}=\mathbf{0}$ and $\iota_h = \iota_h(s_h,a_h)$.
\end{proof}

\begin{lemma}[Decomposition of Suboptimality (DRO version)]
    \label{lemma:decomposition}
    \begin{align*}
    \begin{aligned}
    &\operatorname{SubOpt}(\widehat{\pi}; \gP) = \sum_{h=1}^{H} \mathbb{E}_{\widehat{\pi}}\left[\iota_{h}\left(s_{h}, a_{h}\right) \mid s_{1}\sim \mu \right]-\sum_{h=1}^{H} \mathbb{E}_{\pi^{*}}\left[\iota_{h}\left(s_{h}, a_{h}\right) \mid s_{1}\sim \mu\right]\\
    &+\sum_{h=1}^{H} \mathbb{E}_{\pi^{*}}\left[\left\langle\widehat{Q}_{h}\left(s_{h}, \cdot\right), \pi_{h}^{*}\left(\cdot \mid s_{h}\right)-\widehat{\pi}_{h}\left(\cdot \mid s_{h}\right)\right\rangle_{\mathcal{A}} \mid s_{1}\sim \mu\right],
    \end{aligned}
    \end{align*}
    where
    \begin{align*}
        \iota_h(s,a) \coloneqq \widehat{Q}_h(s,a) - (r_h(s,a) + \inf_{P_{h+1}\in\sP_{h+1}(\cdot \lvert s,a)}\E_{s^{\prime}\sim P_{h+1}}[\widehat{V}_{h+1}(s^{\prime})]).
    \end{align*}
\end{lemma}

\begin{proof}
By the definition above, the suboptimality of the policy $\hat{\pi}$ can be decomposed as 
\begin{align}
    \label{eq:subopt-decompose}
    \operatorname{SubOpt}(\widehat{\pi};\gP) &= \underbrace{(\E_{s\sim \mu}[V^*_1(s)] - \E_{s\sim \mu}[\widehat{V}_1(s)])}_{\text{\uppercase\expandafter{\romannumeral1}}} + \underbrace{(\E_{s\sim \mu}[\widehat{V}_1(s)] - \E_{s\sim \mu}[V_1^{\widehat{\pi}}(s)])}_{\text{\uppercase\expandafter{\romannumeral2}}},
\end{align}
where $\{\widehat{V}_h\}_{h=1}^H$ are the estimated value functions constructed by any algorithm. 

Apply Lemma~\ref{lem:extended-value-difference} to the I term in~\eqref{eq:subopt-decompose} with $\pi = \widehat{\pi}$, $\pi^{\prime}=\pi^{*}$ and $\{\widehat{Q}_h\}_{h=1}^H$ being the estimated Q-functions constructed by our algorithm~\ref{alg:drvi}, we have
\begin{align}
    \label{eq:hat-star}
    \widehat{V}_1(s) - V_1^{*}(s) = \sum_{h=1}^H \E_{\pi^{*}}[\langle \widehat{Q}_h(s_h,\cdot), \widehat{\pi}_h(\cdot\lvert s_h) - \pi^{*}_h(\cdot\lvert s_h) \rangle \lvert s_1=s] + \sum_{h=1}^H\E_{\pi^{*}}[\iota_h(s_h,a_h)\lvert s_1=s].
\end{align}

Similarily, apply Lemma~\ref{lem:extended-value-difference} to the II term in~\eqref{eq:subopt-decompose} with $\pi = \pi^{\prime}=\widehat{\pi}$ and $\{\widehat{Q}_h\}_{h=1}^H$ being the estimated Q-functions constructed by our algorithm~\ref{alg:drvi}, we have 
\begin{align}
    \label{eq:hat-hat}
    \widehat{V}_1(s) - V_1^{\widehat{\pi}}(s) = \sum_{h=1}^H \E_{\widehat{\pi}}[\iota_h(s_h, a_h)\lvert s_1 = s].
\end{align}
Putting \eqref{eq:hat-star} and \eqref{eq:hat-hat} into \eqref{eq:subopt-decompose} we yield the desired conclusions.
\end{proof}

\begin{proof}[Proof of Lemma~\ref{lemma:span} ]
    \label{proof:span}

    Recall the Bellman equation for the $d$-rectangular robust MDP (Equation~\ref{eq:Bellman}):
    \begin{equation}
        \begin{aligned}
        (\sB_h V)(s,a) &= r(s,a) + \inf_{P_h\in \gP^{\operatorname{KL}}( \psi_h;\rho)}\E_{s^{\prime}\sim P_h(\cdot\lvert s,a)}[V(s^{\prime})]\\
        &= \sum_{i\in[d]}\phi_i(s,a)\theta_{h,i} + \sum_{i\in [d]}\phi_i(s,a)\min_{\psi^{\prime}_{h,i}\in \gP^{\operatorname{KL}}(\psi_{h,i}; \rho)} \psi_{h,i}^{\prime \top}V
        \end{aligned}
    \end{equation}

    Since $M\in \gM^{\operatorname{rob}}$, from the proof of above that for any $f\in \gF$, we have $\sB_h f\in \gF$ for any $h\in [H]$, which finish the second part of the proof of lemma~\ref{lemma:span}.
\end{proof}

\begin{lemma}
    For any fix $h\in[H]$ and $i\in[d]$,
    we denote 
     \begin{align*}
        \beta^{*}_{h,i} \in \argmax_{\beta_{h,i}\ge 0} \{-\beta_{h,i} \cdot \E_{\psi_{h,i}}[e^{-V_{h+1}(s^{\prime})/\beta_{h,i}}] - \beta_{h,i} \rho \}.
     \end{align*}
     Then $\beta^{*}_{h,i}\le \overline{\beta}\coloneqq \frac{H-h+1}{\rho}$.
\end{lemma}
\begin{proof}
    This proof is by invoking the part 2 in the Lemma 4 in \cite{zhou2021finite} with $M=H$.
\end{proof}

Thus in the following, we consider the variant of the dual form of the KL optimization, 
\begin{align*}
    \sup_{\beta_{h,i} \in[0,\overline{\beta}]} \{-\beta_{h,i} \cdot \E_{\psi_{h,i}}[e^{-V_{h+1}(s^{\prime})/\beta_{h,i}}] - \beta_{h,i} \rho \}.
\end{align*}

\section{Proof of Theorem~\ref{thm:model-misspecific-sufficient}}
Before we start the proof,
it is obvious to note that $\sum_{i=1}^d \phi_{i}(s,a) = 1, \forall (s,a)\in \gS\times\gA$ under the Assumption~\ref{as:factor-mdp}.
In this section, we mainly prove the Theorem~\ref{thm:model-misspecific-sufficient}. By setting the model mis-specification $\xi=0$, we can recover the results in Theorem~\ref{thm:sufficient-coverage}.

\begin{proposition}
    \label{prop:iota-ols}
    With probability at least $1-\delta$, for all $h\in [H]$, we have 
    \begin{align*}
        \iota_h(s, a)\le & \underline{\beta}(e^{H/\underline{\beta}}-1)(2\xi\sqrt{d} + 10 \sqrt{d\zeta_1} )\sum_{i=1}^d  \lVert \phi_i(s,a) \mathds{1}_i \rVert_{\Lambda_h^{-1}}+ 2\sqrt{2}\sqrt{\underline{\beta}}(e^{H/\underline{\beta}}-1)\sqrt{H\zeta_2}\sum_{i=1}^d  \lVert \phi_i(s,a) \mathds{1}_i \rVert_{\Lambda_h^{-1}} + (H-h)\xi,
    \end{align*}
    for $\zeta_1 = \zetaI$ and $\zeta_2 = \zetaII$.
\end{proposition}

\begin{proof}
    From the DRO Bellman optimality equation and Lemma~\ref{lem:huKL}, we denote 
    \begin{align*}
        (\sB_h \widehat{V}_{h+1})(s,a) &=  r_h(s,a) + \inf_{P_{h+1}\in \sP_{h+1}} \ E_{P_{h+1}(\cdot \lvert s,a)}[\widehat{V}_{h+1}(\spr)]\\
        &=  r_h(s,a) + \inf_{\tilde{P}_{h+1}\in \tilde{\sP}_{h+1}} \ E_{\tilde{P}_{h+1}(\cdot \lvert s,a)}[\widehat{V}_{h+1}(\spr)] \\
        & \quad + (\inf_{P_{h+1}\in \sP_{h+1}} \ E_{P_{h+1}(\cdot \lvert s,a)}[\widehat{V}_{h+1}(\spr)]-\inf_{\tilde{P}_{h+1}\in \tilde{\sP}_{h+1}} \ E_{\tilde{P}_{h+1}(\cdot \lvert s,a)}[\widehat{V}_{h+1}(\spr)])\\
        &= \sum_{i=1}^d \phi_i(s,a)\theta_{h,i} + \sum_{i=1}^d \phi_i(s,a) \maxb\{-\beta_{h,i} \cdot \log (\E_{\psi_{h,i}}[e^{-\widehat{V}_{h+1}(\spr)/\beta_{h,i}}]) - \beta_{h,i}\rho \} \\
        &\quad + (H-h)\xi.
    \end{align*}
    
    Combined with the empirical Bellman operator in our algorithm~\ref{alg:drvi},
    \begin{equation}
        \label{eq:Bellman-error}
        \begin{aligned}
        \iota_h(s, a) &= (\sB_h \widehat{V}_{h+1})(s,a) -  (\widehat{\sB}_h \widehat{V}_{h+1})(s,a) \\
        & = \phi(s,a)^{\top} (\theta_h - \widehat{\theta}_h) + \phi(s,a)^{\top}(w_h-\widetilde{w}_h) + (H-h)\xi,
        \end{aligned}
    \end{equation}
    where $w_{h,i}\coloneqq \maxb\{-\beta_{h,i} \log(\E_{\psi_{h,i}}[e^{-\widehat{V}_{h+1}(\spr)/\beta_{h,i}}]) - \beta_{h,i} \rho \}$ and \\
    $\widetilde{w}_{h,i}\coloneqq \maxb\{-\beta_{h,i} \log(\widehat{\psi}_{h,i}^{\top}[e^{-\widehat{V}_{h+1}(\spr)/\beta_{h,i}}-1]+1) - \beta_{h,i} \rho \}$.

    \textbf{Step 1:} we analyze the error in the reward estimation, i.e., $\phi(s,a)^{\top} (\theta_h - \widehat{\theta}_h)$.
    \begin{equation}
    \label{eq:error-theta}
    \begin{aligned}
        \phi(s,a)^{\top} (\theta_h - \widehat{\theta}_h) & = \phi(s,a)^{\top}\Lambda_h^{-1}\Lambda_h\theta_h - \phi(s,a)^{\top}\Lambda_h^{-1}[\sum_{\tau=1}^N \phi(s_h^{\tau},a_h^{\tau})r_h(s_{h}^{\tau}, a_{h}^{\tau})]\\
        &=  \phi(s,a)^{\top}\Lambda_h^{-1}\Lambda_h\theta_h - \phi(s,a)^{\top}\Lambda_h^{-1}[\sum_{\tau=1}^N \phi(s_h^{\tau},a_h^{\tau})\phi(s_h^{\tau},a_h^{\tau})^{\top}\theta_h]\\
        &= \phi(s,a)^{\top}\Lambda_h^{-1}\Lambda_h\theta_h - \phi(s,a)^{\top}\Lambda_h^{-1}(\Lambda_h-\lambda I)\theta_h\\
        &= \lambda \phi(s,a)^{\top}\Lambda_h^{-1}\theta_h\\
        &\le \lambda \lVert \theta_h \rVert_{\Lambda_h^{-1}} \lVert \phi(s,a)\rVert_{\Lambda^{-1}_h}\\
        & \le \sqrt{d\lambda} \lVert \phi(s,a)\rVert_{\Lambda_h^{-1}}\\
        &\le \sqrt{d\lambda}\sum_{i=1}^d \lVert \phi_i(s,a)\mathds{1}_i \rVert_{\Lambda_h^{-1}}.
    \end{aligned}
    \end{equation}
    Here the last inequality is from 
    \begin{align*}
        \lVert \theta_h\rVert_{\Lambda_h^{-1}} = \sqrt{\theta_h^{\top}\Lambda_h^{-1}\theta_h}\le \lVert \Lambda_h^{-1}\rVert^{1/2}\lVert \theta_h\rVert \le \sqrt{d/\lambda},
    \end{align*}
    by using the fact that $\lVert \Lambda_h^{-1}\rVert\le \lambda^{-1}$ and the Definition~\ref{def:linear}.
    
    \textbf{Step 2:} we turn to the estimation error from the transition model, i.e., $\phi(s,a)^{\top}(w_h-\widehat{w}_h)$.
    We define two auxiliary functions: $$\hat{g}_{h,i}(\beta)\coloneqq -\beta \cdot \log (\widehat{\psi}_{h,i}^{\top}[e^{-\widehat{V}_{h+1}(\spr)/\beta}-1]+ 1) - \beta\rho,$$ and $$g_{h,i}(\beta) \coloneqq -\beta \cdot \log (\E_{\psi_{h,i}}[e^{-\widehat{V}_{h+1}(\spr)/\beta}]) - \beta\rho.$$ 
    Then
    \begin{align}
        & \lvert \sum_{i\in [d]}\phi_i(s,a) (w_{h,i} - \widetilde{w}_{h,i})\rvert \notag \\
        & \le \sum_{i\in [d]} \lvert \phi_i(s,a) (w_{h,i} - \widetilde{w}_{h,i})\rvert \\
        & =  \sum_{i\in [d]} \phi_i(s,a)\lvert \maxb g_{h,i}(\beta_{h,i})  -  \maxb \hat{g}_{h,i}(\beta_{h,i})\rvert  \notag\\
        &\le \sum_{i\in [d]} \phi_i(s,a)\maxb \lvert g_{h,i}(\beta_{h,i}) - \hat{g}_{h,i}(\beta_{h,i})\rvert \notag \\
        &= \sum_{i\in [d]} \phi_i(s,a)\maxb\{\lvert \beta_{h,i} \cdot ( \log (\widehat{\psi}_{h,i}^{\top}[(e^{-\widehat{V}_{h+1}(\spr)/\beta_{h,i}}-1)]+1)- \log (\E_{\psi_{h,i}}[e^{-\widehat{V}_{h+1}(\spr)/\beta_{h,i}}]) ) \rvert \} \notag \\
        &= \sum_{i\in [d]} \phi_i(s,a)\maxb\{\lvert \beta_{h,i}  \cdot ( \log (\widehat{\psi}_{h,i}^{\top}[e^{(H-\widehat{V}_{h+1}(\spr))/\beta_{h,i}}-e^{H/\beta_{h,i}}]+e^{H/\beta_{h,i}})- \log (\E_{\psi_{h,i}}[e^{(H-\widehat{V}_{h+1}(\spr))/\beta_{h,i}}]) ) \rvert \} \notag \\
        &\le \sum_{i\in [d]} \phi_i(s,a)\maxb\{\lvert  \beta_{h,i} \cdot (\widehat{\psi}_{h,i}^{\top}[e^{(H-\widehat{V}_{h+1}(\spr))/\beta_{h,i}}-e^{H/\beta_{h,i}}] + e^{H/\beta_{h,i}}-  \E_{\psi_{h,i}}[e^{(H-\widehat{V}_{h+1}(\spr))/\beta_{h,i}}])\rvert\}\label{eq:w_difference}, 
    \end{align}
where the last inequality follows from the fact~\ref{fa:exp-2} by setting $x=\beta_{h,i}  \log (\widehat{\psi}_{h,i}^{\top}[e^{(H-\widehat{V}_{h+1}(s))/\beta_{h,i}} - e^{H/\beta_{h,i}}] + e^{H/\beta_{h,i}})$ and $y=\beta_{h,i}  \log \E_{\psi_{h,i}}[e^{(H-\widehat{V}_{h+1}(s))/\beta_{h,i}}]$. 
To ease the presentation, we index the finite states from $1$ to $S$ and introduce the vector $\widehat{V}_{h+1}\in\R^{S}$ where $[\widehat{V}_{h+1}]_{j} = \widehat{V}_{h,i}(s_j)$ and $[H-\widehat{V}_{h+1}]_j = H-\widehat{V}_{h+1}(s_j)$.


We denote $\mathds{1}_j\in \R^{d\times 1}$ with the $j$-the component as 1 and the other components are 0 and $\bm{1}\in \R^{S\time 1}$ is a all-one vector.
Notice that 
\begin{align*}
    & \E_{\psi_{h,i}}[e^{(H-\widehat{V}_{h+1}(\spr))/\beta_{h,i}}] \\
    &= \E_{\psi_{h,i}}[e^{(H-\widehat{V}_{h+1}(\spr))/\beta_{h,i}}-e^{H/\beta_{h,i}}] + e^{H/\beta_{h,i}}\\ 
    &= \psi_{h,i}^{\top}(e^{(H-\widehat{V}_{h+1})/\beta_{h,i}}-e^{H/\beta_{h,i}}) + e^{H/\beta_{h,i}}\\
    &= \mathds{1}_i^\top \psi_{h}^{\top} (e^{(H-\widehat{V}_{h+1})/\beta_{h,i}}-e^{H/\beta_{h,i}})+e^{H/\beta_{h,i}}\\
    &= \mathds{1}_i^\top \Lambda_h^{-1}\Lambda_h\psi_{h}^{\top}(e^{(H-\widehat{V}_{h+1})/\beta_{h,i}}-e^{H/\beta_{h,i}}) + e^{H/\beta_{h,i}}\\
    &= \mathds{1}_i^\top \Lambda_h^{-1}(\sum_{\tau=1}^N \phi(s_{h}^{\tau}, a_{h}^{\tau})\phi(s_{h}^{\tau}, a_{h}^{\tau})^{\top}+\lambda I)\psi_{h}^{\top}(e^{(H-\widehat{V}_{h+1})/\beta_{h,i}}-e^{H/\beta_{h,i}}) + e^{H/\beta_{h,i}}\\
    &= \mathds{1}_i^\top \Lambda_h^{-1}\sum_{\tau=1}^N \phi(s_{h}^{\tau}, a_{h}^{\tau})\phi(s_{h}^{\tau}, a_{h}^{\tau})^{\top}\psi_{h}^{\top}(e^{(H-\widehat{V}_{h+1})/\beta_{h,i}}-e^{H/\beta_{h,i}})\\
    &\quad +\lambda \mathds{1}_i^\top \Lambda_h^{-1}\psi^{\top}_{h} (e^{(H-\widehat{V}_{h+1})/\beta_{h,i}}-e^{H/\beta_{h,i}}) + e^{H/\beta_{h,i}}\\
    &= \mathds{1}_i^\top \Lambda_h^{-1}\sum_{\tau=1}^N \phi(s_{h}^{\tau}, a_{h}^{\tau})\tilde{P}_{h+1}(\cdot\lvert s_{h}^{\tau}, a_{h}^{\tau})^{\top}(e^{(H-\widehat{V}_{h+1})/\beta_{h,i}}-e^{H/\beta_{h,i}})\\
    &\quad +\lambda \mathds{1}_i^\top\Lambda_h^{-1}\psi^{\top}_{h} (e^{(H-\widehat{V}_{h+1})/\beta_{h,i}}-e^{H/\beta_{h,i}}) + e^{H/\beta_{h,i}},
\end{align*}
and 
\begin{align*}
    & \widehat{\psi}_{h,i}^{\top}[(e^{(H-\widehat{V}_{h+1}(s^{\prime}))/\beta_{h,i}}-e^{H/\beta_{h,i}})] + e^{H/\beta_{h,i}}\\
    &= \widehat{\psi}_{h,i}^{\top}(e^{(H-\widehat{V}_{h+1})/\beta_{h,i}}-e^{H/\beta_{h,i}}) + e^{H/\beta_{h,i}}\\
    &=\mathds{1}_i^\top\Lambda_h^{-1}\sum_{\tau=1}^N \phi(s_h^{\tau}, a_h^{\tau})\mathds{1}(s_{h+1}^{\tau})^{\top}(e^{(H-\widehat{V}_{h+1})/\beta_{h,i}}-e^{H/\beta_{h,i}}) + e^{H/\beta_{h,i}},
\end{align*}
where $\mathds{1}(s_{h+1}^{\tau})\in \R^{d\times 1}$ with the correponding component for the $s_{h+1}^{\tau}$ being 1 and the other being 0 and  $\psi_{h}=[\psi_{h,1}, \psi_{h,2},\cdots, \psi_{h,d}]\in\R^{d\times S}.$

We consider their difference,
\begin{align*}
    & \quad \left \lvert \beta_{h,i} \left(\widehat{\psi}_{h,i}^{\top}[(e^{(H-\widehat{V}_{h+1}(\spr))/\beta_{h,i}}-e^{H/\beta_{h,i}})] + e^{H/\beta_{h,i}} - \E_{\psi_{h,i}}[e^{(H-\widehat{V}_{h+1}(\spr))/\beta_{h,i}}]\right)\right \rvert\\
    &= \left \lvert \beta_{h,i} \mathds{1}_i^\top\Lambda_h^{-1}\left (\sum_{\tau=1}^N \phi(s_{h}^{\tau}, a_{h}^{\tau})\cdot (\mathds{1}(s_{h+1}^{\tau}) - \tilde{P}_{h+1}(\cdot\lvert s_{h}^{\tau}, a_{h}^{\tau}))^{\top}(e^{(H-\widehat{V}_{h+1})/\beta_{h,i}}-e^{H/\beta_{h,i}})\right) \right \rvert \\
    &\quad + \left \lvert \lambda \beta_{h,i} \mathds{1}_i^\top \Lambda_h^{-1}\psi_{h}^{\top} (e^{(H-\widehat{V}_{h+1})/\beta_{h,i}}-e^{H/\beta_{h,i}}) \right \rvert\\
    &= \left \lvert \beta_{h,i} \mathds{1}_i^\top\Lambda_h^{-1}\left (\sum_{\tau=1}^N \phi(s_{h}^{\tau}, a_{h}^{\tau})\cdot (P_{h+1}(\cdot\lvert s_{h}^{\tau}, a_{h}^{\tau}) - \tilde{P}_{h+1}(\cdot\lvert s_{h}^{\tau}, a_{h}^{\tau}))^{\top}(e^{(H-\widehat{V}_{h+1})/\beta_{h,i}}-e^{H/\beta_{h,i}})\right) \right \rvert\\
    &\quad + \left \lvert \beta_{h,i} \mathds{1}_i^\top\Lambda_h^{-1}\left (\sum_{\tau=1}^N \phi(s_{h}^{\tau}, a_{h}^{\tau})\cdot(\mathds{1}(s_{h+1}^{\tau}) - P_{h+1}(\cdot\lvert s_{h}^{\tau}, a_{h}^{\tau}))^{\top}(e^{(H-\widehat{V}_{h+1})/\beta_{h,i}}-e^{H/\beta_{h,i}})\right) \right \rvert\\
    &\quad + \left \lvert \lambda \beta_{h,i} \mathds{1}_i^\top \Lambda_h^{-1}\psi_{h}^{\top} (e^{(H-\widehat{V}_{h+1})/\beta_{h,i}}-e^{H/\beta_{h,i}})\right \rvert\\
    &\le \left \lvert \xi \beta_{h,i} (e^{H/\beta_{h,i}}-1) \cdot \mathds{1}_i^\top\Lambda_h^{-1}\sum_{\tau=1}^N \phi(s_{h}^{\tau}, a_{h}^{\tau}) + \beta_{h,i} \mathds{1}_i^\top\Lambda_h^{-1}\left (\sum_{\tau=1}^N \phi(s_{h}^{\tau}, a_{h}^{\tau})\cdot \eps_{h}^{\tau}(\beta_{h,i}, \widehat{V}_{h+1})\right)\right \rvert\\
    &\quad + \left \lvert \lambda \beta_{h,i} \mathds{1}_i^\top\Lambda_h^{-1}\psi_{h}^{\top} (e^{(H-\widehat{V}_{h+1})/\beta_{h,i}}-e^{H/\beta_{h,i}}) \right \rvert,
\end{align*}
where $\epsilon_h^{\tau}(\beta, V) \coloneqq (P_{h+1}(\cdot\lvert s_{h}^{\tau}, a_{h}^{\tau})-\mathds{1}(s_{h+1}^{\tau}))^{\top}(e^{(H-V)/\beta_{h,i}}-e^{H/\beta_{h,i}})$ and $\lvert e^{(H-V)/\beta_{h,i}}-e^{H/\beta_{h,i}}\rvert\le (e^{H/\beta_{h,i}}-1)$.

Plug into Equation~\ref{eq:w_difference} we have 
\begin{equation}
    \label{eq:decomposition}
\begin{aligned}
    & \lvert \phi(s,a)^{\top}(w_h-\widetilde{w}_h)\rvert\\
    & = \lvert \sum_{i=1}^d \phi_i(s,a) (w_{h,i}-\widetilde{w}_{h,i})\rvert\\
    & \le \underbrace{\xi \sum_{i=1}^d\max_{\beta_{h,i}\in[\underline{\beta},\overline{\beta}]} \phi_i(s,a)\lvert \beta_{h,i} (e^{H/\beta_{h,i}}-1)  \cdot \mathds{1}_i^\top\Lambda_h^{-1}\sum_{\tau=1}^N \phi(s_{h}^{\tau}, a_{h}^{\tau}) \rvert}_{\operatorname{I}} \\
    & + \underbrace{\sum_{i=1}^d \phi_i(s,a) \lvert  \mathds{1}_i^\top\Lambda_h^{-1}(\sum_{\tau=1}^N \phi(s_{h}^{\tau}, a_{h}^{\tau})\cdot \max_{\beta_{h,i}\in[\underline{\beta}, \overline{\beta}]}\beta_{h,i} \eps_{h}^{\tau}(\beta_{h,i}, \widehat{V}_{h+1}))\rvert}_{\operatorname{II}}\\
    & + \underbrace{\sum_{i=1}^{d} \max_{\beta_{h,i}\in[\underline{\beta},\overline{\beta}]} \phi_i(s,a) \lvert \lambda  \beta_{h,i} \mathds{1}_i^\top\Lambda_h^{-1}\psi_{h}^{\top} (e^{(H-\widehat{V}_{h+1})/\beta_{h,i}}-e^{H/\beta_{h,i}}) \big) \rvert}_{\operatorname{III}}.
\end{aligned}
\end{equation}

For $\operatorname{I}$ term 
in \ref{eq:decomposition}, for any $\beta_{h,i}\in[\underline{\beta},\overline{\beta}]$ we know,
\begin{align*}
    &\xi \sum_{i=1}^d  \lvert \beta_{h,i} (e^{H/\beta_{h,i}}-1)   \cdot \phi_i(s,a) \mathds{1}_i^\top\Lambda_h^{-1}\sum_{\tau=1}^N \phi(s_{h}^{\tau}, a_{h}^{\tau})  \rvert\\
    &\le \xi \sum_{i=1}^d  \lVert \beta_{h,i} (e^{H/\beta_{h,i}}-1)   \phi_i(s,a) \cdot \mathds{1}_i \rVert_{\Lambda_h^{-1}} \lVert \sum_{\tau=1}^N \phi(s_{h}^{\tau}, a_{h}^{\tau}) \rVert_{\Lambda_h^{-1}}.
\end{align*}

Note that 
\begin{align*}
    \sum_{i=1}^d \lVert \beta_{h,i} (e^{H/\beta_{h,i}}-1)  \phi_i(s,a)\cdot \mathds{1}_i \rVert_{\Lambda_h^{-1}} 
    &\le \underline{\beta}(e^{H/\underline{\beta}} - 1)\sum_{i=1}^d \lVert \phi_i(s,a)\mathds{1}_i \rVert_{\Lambda_h^{-1}}.
\end{align*}

\begin{align*}
    \lVert \sum_{\tau=1}^{N}\phi(s_h^{\tau}, a_h^{\tau})\rVert_{\Lambda_h^{-1}}&= \sqrt{(\sum_{\tau=1}^{N}\phi(s_h^{\tau}, a_h^{\tau}))^{\top} \Lambda_h^{-1}(\sum_{\tau=1}^{N}\phi(s_h^{\tau}, a_h^{\tau}))} \\
    &= \sqrt{\operatorname{Tr}(\Lambda_h^{-1}(\sum_{\tau=1}^{N}\phi(s_h^{\tau}, a_h^{\tau}))(\sum_{\tau=1}^{N}\phi(s_h^{\tau}, a_h^{\tau}))^{\top})}\\
    &= \sqrt{\operatorname{Tr}(\Lambda_h^{-1}(\Lambda_h - \lambda \cdot I))}\\
    &\le \sqrt{\operatorname{Tr}(\Lambda_h^{-1} \Lambda_h)}\\
    & = \sqrt{d}.
\end{align*}

Thus 
\begin{align*}
    \operatorname{I} \le  \xi \sqrt{d} \underline{\beta}(e^{H/\underline{\beta}} - 1)\sum_{i=1}^d \lVert \phi_i(s,a)\mathds{1}_i \rVert_{\Lambda_h^{-1}}.
\end{align*}

For term $\operatorname{III}$ and any $\beta_{h,i}\in [\underline{\beta}, \overline{\beta}]$,
\begin{align*}
    &\sum_{i=1}^{d} \lvert \lambda\beta_{h,i} \phi_i(s,a)\mathds{1}_i^\top\Lambda_h^{-1}\psi_{h}^{\top} (e^{(H-\widehat{V}_{h+1})/\beta_{h,i}}-e^{H/\beta_{h,i}}) \rvert\\
    &\le  \sum_{i=1}^{d} \lambda \lVert  \phi_i(s,a) \mathds{1}_i^\top\Lambda_h^{-1} \rVert_1 \lVert \beta_{h,i} \psi_{h}^{\top}  (e^{(H-\widehat{V}_{h+1})/\beta_{h,i}}-e^{H/\beta_{h,i}})  \rVert_{\infty}\\
    &\le \lambda \underline{\beta}(e^{H/\underline{\beta}} - 1) \sum_{i=1}^{d} \lVert \phi_i(s,a)\mathds{1}_i^\top\Lambda_h^{-1} \rVert_{1}\\
    &\le \sqrt{d}\lambda \underline{\beta}(e^{H/\underline{\beta}} - 1) \sum_{i=1}^{d} \lVert \phi_i(s,a) \mathds{1}_i^\top\Lambda_h^{-1} \rVert_{2}\\
    &\le \sqrt{d}\lambda \underline{\beta}(e^{H/\underline{\beta}} - 1) \lVert\Lambda_h^{-1/2} \rVert_{2} \sum_{i=1}^{d} \lVert  \phi_i(s,a) \mathds{1}_i \rVert_{\Lambda_h^{-1}}\\    
    &\le \sqrt{d \lambda} \underline{\beta}(e^{H/\underline{\beta}} - 1)\sum_{i=1}^{d} \lVert  \phi_i(s,a) \mathds{1}_i \rVert_{\Lambda_h^{-1}},
\end{align*}
the second inequality is from $\lVert e^{(H-\widehat{V}_{h+1})/\beta_{h,i}}-e^{H/\beta_{h,i}}\rVert_{\infty}\le (e^{H/\underline{\beta}} - 1)$ and the three inequality is from $\lVert x\rVert_{1}\le \sqrt{d}\lVert x\rVert_{2}$ for any $x\in \sR^{d}$.

To control $\operatorname{II}$ term, we invoke Lemma~\ref{lem:concentration-our} with the choice of $\lambda=1$ and then with probability at least $1-\delta$, 
\begin{align*}
    &\sum_{i=1}^d \lvert \phi_i(s,a) \mathds{1}_i^\top\Lambda_h^{-1}(\sum_{\tau=1}^N \phi(s_{h}^{\tau}, a_{h}^{\tau})\cdot \max_{\beta_{h,i}\in[\underline{\beta}, \overline{\beta}]}\beta_{h,i} \eps_{h}^{\tau}(\beta_{h,i}, \widehat{V}_{h+1})) \rvert\\
    &\le \sum_{i=1}^d  \lVert \phi_i(s,a) \mathds{1}_i \rVert_{\Lambda_h^{-1}} \lVert \sum_{\tau=1}^N \phi(s_{h}^{\tau}, a_{h}^{\tau})\cdot \max_{\beta_{h,i}\in[\underline{\beta}, \overline{\beta}]}\beta_{h,i} \eps_{h}^{\tau}(\beta_{h,i}, \widehat{V}_{h+1}) \rVert_{\Lambda_h^{-1}}\\
    & \le (4\sqrt{d}\underline{\beta}(e^{H/\underline{\beta}}-1) \sqrt{\zetaI}\\
    &\quad + 2\sqrt{2}\underline{\beta}(e^{H/\underline{\beta}} - 1) (\sqrt{\frac{H}{\underline{\beta}} \cdot \zetaII}
    +\sqrt{2}))\sum_{i=1}^d  \lVert \phi_i(s,a) \mathds{1}_i \rVert_{\Lambda_h^{-1}}\\
    &\le (e^{H/\underline{\beta}}-1)(8\underline{\beta}\sqrt{d\zeta_1}+ 2\sqrt{2} \sqrt{\underline{\beta}H\zeta_2})\sum_{i=1}^d  \lVert \phi_i(s,a) \mathds{1}_i \rVert_{\Lambda_h^{-1}},
\end{align*}
where the first inequality is from $\sqrt{x+y}\le \sqrt{x} + \sqrt{y}$ for $x,y\ge 0$.
The second inequality is by $\sqrt{d\zeta_1}\ge 1$ where $\zeta_1 = \zetaI$ and $\zeta_2=\zetaII$.
Thus we have 
\begin{equation}
\label{eq:error-w}
\begin{aligned}
    &\lvert \phi(s,a)^{\top}(w_h-\widetilde{w}_h)\rvert\\
    &\le \underline{\beta}(e^{H/\underline{\beta}}-1)(\xi\sqrt{d}  + 8 \cdot \sqrt{d\zeta_1} + \sqrt{d})\sum_{i=1}^d  \lVert \phi_i(s,a) \mathds{1}_i \rVert_{\Lambda_h^{-1}} + 2\sqrt{2} \sqrt{\underline{\beta}}(e^{H/\underline{\beta}}-1)\sqrt{H\zeta_2}\sum_{i=1}^d  \lVert \phi_i(s,a) \mathds{1}_i \rVert_{\Lambda_h^{-1}}\\
    &\quad + \xi \sqrt{d} \underline{\beta}(e^{H/\underline{\beta}} - 1)\sum_{i=1}^d \lVert \phi_i(s,a)\mathds{1}_i \rVert_{\Lambda_h^{-1}}\\
    &\le \underline{\beta}(e^{H/\underline{\beta}}-1)(\xi\sqrt{d}  + 9 \cdot \sqrt{d\zeta_1})\sum_{i=1}^d  \lVert \phi_i(s,a) \mathds{1}_i \rVert_{\Lambda_h^{-1}} + 2\sqrt{2} \sqrt{\underline{\beta}}(e^{H/\underline{\beta}}-1)\sqrt{H\zeta_2}\sum_{i=1}^d  \lVert \phi_i(s,a) \mathds{1}_i \rVert_{\Lambda_h^{-1}}\\
    & \quad +\xi \sqrt{d} \underline{\beta}(e^{H/\underline{\beta}} - 1)\sum_{i=1}^d \lVert \phi_i(s,a)\mathds{1}_i \rVert_{\Lambda_h^{-1}},
\end{aligned}
\end{equation}
where the second inequality is by noticing that $f(\beta) = \beta (e^{H/\beta}-1)$ and $g(\beta) = \sqrt{\beta}(e^{H/\beta}-1)$ are both monotonically decreasing with $\beta>0$ and the last inequality is from $\sqrt{d\zeta_1}\ge 1$.

Plug Equation~\ref{eq:error-theta} and ~\ref{eq:error-w} into Equation~\ref{eq:Bellman-error} to finally upper bound the Bellman error,
\begin{align*}
    \lvert \iota_h(s,a)\rvert& = \lvert (\sB_h \widehat{V}_{h+1})(s,a) -  (\hat{\sB}_h \widehat{V}_{h+1})(s,a)\rvert\\
    & \le \sqrt{d} \sum_{i=1}^d  \lVert \phi_i(s,a) \mathds{1}_i \rVert_{\Lambda_h^{-1}} + \underline{\beta}(e^{H/\underline{\beta}}-1)(\xi\sqrt{d}  + 9 \cdot \sqrt{d\zeta_1})\sum_{i=1}^d  \lVert \phi_i(s,a) \mathds{1}_i \rVert_{\Lambda_h^{-1}}\\
    &\quad + 2\sqrt{2} \sqrt{\underline{\beta}}(e^{H/\underline{\beta}}-1)\sqrt{H\zeta_2}\sum_{i=1}^d  \lVert \phi_i(s,a) \mathds{1}_i \rVert_{\Lambda_h^{-1}} + \xi \sqrt{d} \underline{\beta}(e^{H/\underline{\beta}} - 1)\sum_{i=1}^d \lVert \phi_i(s,a)\mathds{1}_i \rVert_{\Lambda_h^{-1}} + (H-h)\xi\\
    &\le \underline{\beta}(e^{H/\underline{\beta}}-1)(2\xi\sqrt{d} + 10 \sqrt{d\zeta_1} )\sum_{i=1}^d  \lVert \phi_i(s,a) \mathds{1}_i \rVert_{\Lambda_h^{-1}}+ 2\sqrt{2}\sqrt{\underline{\beta}}(e^{H/\underline{\beta}}-1)\sqrt{H\zeta_2}\sum_{i=1}^d  \lVert \phi_i(s,a) \mathds{1}_i \rVert_{\Lambda_h^{-1}}\\
    &\quad + (H-h)\xi,
\end{align*}
where the last inequality is from the fact that $f(\beta) = \beta(e^{H/\beta}-1)\ge H\ge 1$ and and $\sqrt{d \zeta_1}>1$.
\end{proof}

\begin{proof}[Proof of Theorem~\ref{thm:model-misspecific-sufficient}]
Our policy is the greedy policy w.r.t. to the estimated Q-function, thus we can reduce the suboptimality reduction in Lemma~\ref{lemma:decomposition} into
\begin{align}
    \label{eq:sub-opt-1}
    \operatorname{SubOpt}(\widehat{\pi} ; \gP)\le \sum_{h=1}^{H} \mathbb{E}_{\widehat{\pi}}\left[\iota_{h}\left(s_{h}, a_{h}\right) \mid s_{1}\sim \mu \right]-\sum_{h=1}^{H} \mathbb{E}_{\pi^{*}}\left[\iota_{h}\left(s_{h}, a_{h}\right) \mid s_{1}\sim \mu\right].
\end{align}
    Putting Proposition~\ref{prop:iota-ols} in the Equation~\ref{eq:sub-opt-1} we know that with probability at least $1-\delta/2$,
    \begin{align*}
        & \operatorname{SubOpt}(\widehat{\pi};\gP)\\
        &\le (e^{H/\underline{\beta}}-1)(2\xi\sqrt{d}\underline{\beta} + 10\cdot \sqrt{d\zeta_1}\underline{\beta}+ 2\sqrt{2} \sqrt{\underline{\beta}}\sqrt{H\zeta_2}) \cdot \sum_{h=1}^H (\E_{\widehat{\pi}}[\lVert \Lambda_h^{-1}\rVert_{\operatorname{tr}(\phi(s,a))}]+ \E_{\pi^{*}}[\lVert \Lambda_h^{-1}\rVert_{\operatorname{tr}(\phi(s,a))}])+\cdots\\
        & \quad + (H-1)H\xi/2.
    \end{align*}
    Based on the Assumption~\ref{as:partial-coverage}, Definition~\ref{def:linear} and using the similar steps in the proof of Corollar 4.6 in \cite{jin2021pessimism}, we can conclude that when $N$ is sufficiently large so that $N\ge 40/\underline{c}\cdot \log(4dH/\delta)$, for all $h\in[H]$, it holds that with probability at least $1-\delta/2$,
    \begin{align*}
        \lVert \Lambda_h^{-1}\rVert_{\operatorname{tr}(\phi(s,a))} &= \sum_{i=1}^d \lVert \phi_i(s,a)\mathds{1}_i \rVert_{\Lambda_h^{-1}} \\
        &\le \sum_{i=1}^d  \lVert \phi_i(s,a)\mathds{1}_i\rVert \lVert\Lambda_h^{-1}\rVert^{-1/2}\\
        &\le \sqrt{\frac{2}{\underline{c}N}}\coloneqq c/\sqrt{N}, \quad \forall (s,a)\in \gS\times\gA, \forall h\in [H],
    \end{align*}
    where the second inequality is from Assumption~\ref{as:factor-mdp}.
    In conclusions, when $N\ge 40/\underline{c}\cdot \log(4dH/\delta)$, we have probability at least $1-\delta$, 
    \begin{align*}
        \operatorname{SubOpt}(\widehat{\pi};\gP)&\le c_1 H\underline{\beta}(e^{H/\underline{\beta}}-1)(\xi\sqrt{d} + \sqrt{d\zeta_1})/N^{1/2} + c_2 \sqrt{\underline{\beta}}(e^{H/\underline{\beta}}-1)\sqrt{\zeta_2}H^{3/2}/N^{1/2}\\
        &\quad + (H-1)H\xi,
    \end{align*}
    for some absolute constants $c_1$ and $c_2$ that only depend on $\underline{c}$.
\end{proof}

\section{Auxiliary Lemmas For the Proof for Theorem~\ref{thm:model-misspecific-sufficient}}
\label{subsec:auxiliary}
\begin{lemma}
    \label{lem:concentration-our}
    For all $i\in[d]$, $\beta_{h,i}\in[\underline{\beta},\overline{\beta}]$, 
    and the estimator $\{\widehat{V}_{h}\}_{h=1}^{H}$ constructed from Algorithm~\ref{alg:drvi},
    we have the following holds with probability at least $1-\delta$,
    \begin{align*}
        \lVert \sum_{\tau=1}^N \phi(s_{h}^{\tau}, a_{h}^{\tau})\cdot \eps_h^{\tau}(\beta_{h,i}, \widehat{V}_{h+1})\rVert_{\Lambda_h^{-1}}^2\le & 16d(e^{H/\beta_{h,i}}-1)^2\zeta_1 + 8(e^{H/\beta_{h,i}} - 1)^2(\frac{H}{\beta_{h,i}} \zeta_2 + 2),
    \end{align*}
    for $\zeta_1 = \zetaI$, $\zeta_2 = \zetaII$ and some absolute constant $c>1$.
\end{lemma}

\begin{proof}
For the fixed $h\in[H]$ and fixed $\tau\in[N]$, we define the $\sigma$-algebra,
\begin{align*}
    \mathcal{F}_{h}^{\tau}\coloneqq \sigma(\{s_h^{\tau^{\prime}}, a_h^{\tau^{\prime}}\}_{\tau^{\prime}=1}^{\tau}\cup \{r_h^{\tau^{\prime}}, s_{h+1}^{\tau^{\prime}}\}_{\tau^{\prime}=1}^{(\tau-1)\vee  0}),
\end{align*}
i.e., $\mathcal{F}_{h}^{\tau}$ is the filtration generated by the samples $\{s_h^{\tau}, a_h^{\tau}\}_{\tau^{\prime}=1}^{\tau}\cup \{r_h^{\tau}, s_{h+1}^{\tau}\}_{\tau^{\prime}=1}^{(\tau-1)\vee 0}$. 
Notice that $\E[\epsilon_h^{\tau}(\beta_{h,i}, \widehat{V}_{h+1})\lvert \gF_h^{\tau}]=0$.
However, $\widehat{V}_{h+1}$ depends on $\{(s_{h}^{\tau}, a_{h}^{\tau})\}_{\tau\in [N]}$ via $\{(s_{h^{\prime}}^{\tau}, a_{h^{\prime}}^{\tau})\}_{h^{\prime}>h, \tau\in [N]}$ and thus we cannot directly apply vanilla concentration bounds to control $\lVert \sum_{\tau=1}^N \phi(s_{h}^{\tau}, a_{h}^{\tau})\cdot \eps_h^{\tau}(\beta_{h,i}, \widehat{V}_{h+1})\rVert_{\Lambda_h^{-1}}$. 

To tackle this point, we consider the function family $V_h(R)$. In specific, 
\begin{align*}
    V_h(R) = \{V_h(x;\tilde{w}): \mathcal{S}\rightarrow [0, H-h+1] \lvert \lVert \tilde{w} \rVert \le R\},
\end{align*}
and 
\begin{align*}
V_h(x;\tilde{w})&=\max_{a\in \mathcal{A}}\{\phi(s,a)^{\top}\tilde{w}\}.
\end{align*}

We let $\mathcal{N}_{\epsilon}(R)$ be the minimal $\epsilon$-cover of $\mathcal{V}_h(R)$ with respect to the supremum norm, i.e., for any function $V\in \mathcal{V}_h(R)$, there exists a function $\epsV\in \mathcal{N}_{\epsilon}(R)$ such that 
\begin{align*}
    \sup_{s\in \mathcal{S}}\lvert V(s) - \epsV(s) \rvert\le \epsilon.
\end{align*}

Hence, for $\widehat{V}_{h+1}$,
we have $V^{\dagger}_{h+1}\in \mathcal{N}_{\epsilon}(R)$ such that 
\begin{align*}
    \sup_{s\in \mathcal{S}} \lvert \widehat{V}_{h+1}(s) - V^{\dagger}_{h+1}(s)\rvert \le s.
\end{align*}


For the ease of presentation, we ignore the subscript of $\beta_{h,i}$ and use $\beta$ in the following.
Next we denote $\betaeps$ the minimal $\epsilon$-cover of the $[\underline{\beta}, \overline{\beta}]$ with respect to the absolute value, i.e., for any $\beta \in [\underline{\beta},\overline{\beta}]$, there exists $\epsN(\beta)\in \betaeps$ such that 
\begin{align*}
    \lvert \beta - \epsN(\beta)\rvert \le \epsilon.
\end{align*}

We proceed our analysis for all $i\in[d]$ and $\beta\in [\underline{\beta}, \overline{\beta}]$
\begin{subequations}
\begin{align}
    &  \lVert \sum_{\tau\in[N]} \phi(s_h^{\tau}, a_{h}^{\tau})\epsilon_h^{\tau}(\beta, \widehat{V}_{h+1})\rVert_{\Lambda^{-1}_h}^2 \nonumber\\
    &\le  2\lVert \sum_{\tau\in[N]} \phi(s_h^{\tau}, a_{h}^{\tau})\epsilon_h^{\tau}(\beta, V^{\dagger}_{h+1})\rVert_{\Lambda^{-1}_h}^2 +  
    2\lVert \sum_{\tau\in[N]} \phi(s_h^{\tau}, a_{h}^{\tau})(\epsilon_h^{\tau}(\beta, \widehat{V}_{h+1}) - \epsilon_h^{\tau}(\beta, V^{\dagger}_{h+1})) \rVert_{\Lambda^{-1}_h}^2 \nonumber\\
    & \le \sup_{V\in \mathcal{N}_{\epsilon}(R)} 2\lVert \sum_{\tau\in[N]} \phi(s_h^{\tau}, a_{h}^{\tau})\epsilon_h^{\tau}(\beta, V)\rVert_{\Lambda^{-1}_h}^2 +  
    2\lVert \sum_{\tau\in[N]} \phi(s_h^{\tau}, a_{h}^{\tau})(\epsilon_h^{\tau}(\beta, \widehat{V}_{h+1}) - \epsilon_h^{\tau}(\beta, V^{\dagger}_{h+1})) \rVert_{\Lambda^{-1}_h}^2\nonumber\\
    &\le \underbrace{\sup_{V\in \mathcal{N}_{\epsilon}(R)} 4\lVert \sum_{\tau\in[N]} \phi(s_h^{\tau}, a_{h}^{\tau})\epsilon_h^{\tau}(N_{\epsilon}(\beta), V)\rVert_{\Lambda^{-1}_h}^2}_{\operatorname{I}} + \cdots\nonumber\\
    &\quad + \underbrace{\sup_{V\in \mathcal{N}_{\epsilon}(R)} 4\lVert \sum_{\tau\in[N]} ( \phi(s_h^{\tau}, a_{h}^{\tau})(\epsilon_h^{\tau}(\beta, V) - \epsilon_h^{\tau}(N_\epsilon(\beta), V)) ) \rVert_{\Lambda^{-1}_h}^2}_{\operatorname{II}}+\cdots\nonumber\\
    & \quad + \underbrace{ 
    2\lVert \sum_{\tau\in[N]} \phi(s_h^{\tau}, a_{h}^{\tau})(\epsilon_h^{\tau}(\beta, \widehat{V}_{h+1}) - \epsilon_h^{\tau}(\beta, V^{\dagger}_{h+1})) \rVert_{\Lambda^{-1}_h}^2}_{\operatorname{III}}, \label{eq:decompose-martingale}
\end{align}
\end{subequations}
where the first and second inequality is from the fact that $\lVert a + b\rVert_{\Lambda}^2\le 2\lVert a\rVert_{\Lambda}^2+2\lVert b\rVert_{\Lambda}^2$ for any vectors $a,b\in \R^d$ and any positive definite matrix $\Lambda\in \R^{d\times d}$.

We invoke Lemma~\ref{lem:concentration}, Lemma~\ref{lm:difference-beta} and Lemma~\ref{lm:difference-V} for the term I, II and III respectively and have with probability at least $1-\delta$, for all $h\in[H]$,
\begin{align*}
    & \lVert \sum_{\tau\in[N]} \phi(s_h^{\tau}, a_{h}^{\tau})\epsilon_h^{\tau}(\beta, \widehat{V}_{h+1})\rVert_{\Lambda^{-1}_h}^2 \\
    &\le 4d\left(e^{H / \beta}-1\right)^2 \log \left(1+\frac{N}{\lambda}\right)\\
    & +8\left(e^{H / \beta}-1\right)^2 \cdot \log \left(\frac{H\left|\mathcal{N}_\epsilon(R, B, \lambda)\right|\left|\mathcal{N}_\epsilon(\underline{\beta}, \bar{\beta})\right|}{\delta}\right) \\
    & \quad + \frac{16N^2(H-h)^2\epsilon^2}{\lambda\beta^4}e^{2H/\beta} + \frac{8N^2\epsilon^2}{\lambda\beta^2}e^{2H/\beta}\\
    &\le 4d(e^{H/\beta} - 1)^2\log(1+\frac{N}{\lambda}) + 8(e^{H/\beta} - 1)^2\cdot \log(\frac{H^2}{\epsilon \delta \rho}) + 8 d(e^{H/\beta} - 1)^2\cdot \log(1+\frac{4R}{\epsilon})\\
    &\quad + \frac{16N^2(H-h)^2\epsilon^2}{\lambda\beta^4}e^{2H/\beta} + \frac{8N^2\epsilon^2}{\lambda\beta^2}e^{2H/\beta},
\end{align*}
where the second inequality is from Lemma~\ref{lem:eps-covering} and the fact that $\epsN(\underline{\beta}, \overline{\beta}) \le \frac{H}{\rho\epsilon}$.

By choosing $R=2Hd^{3/2}$, 
$\lambda = 1$, $\epsilon = \frac{2}{4NHe^{H/\underline{\beta}}}$,
then for all $\beta\in [\underline{\beta}, \overline{\beta}]$ 
\begin{align*}
    &\lVert \sum_{\tau\in[N]} \phi(s_h^{\tau}, a_{h}^{\tau})\epsilon_h^{\tau}(\beta, \widehat{V}_{h+1})\rVert_{\Lambda^{-1}_h}^2 \\
    &\le 4d(e^{H/\beta} - 1)^2\cdot \log(2N) 
    + 8(e^{H/\beta} - 1)^2\cdot \frac{H}{\beta} \zetaII
    + \frac{1}{\beta^4} 
    + \frac{1}{2H^2\beta^2}\\
    & \quad 
    + 8d(e^{H/\beta} - 1)^2 \cdot \log(1+16Nd^{3/2}H^2e^{\frac{H}{\underline{\beta}}})
    \\
    &\le 16d(e^{H/\beta}-1)^2 \zetaI\\
    &\quad + 8(e^{H/\beta} - 1)^2\cdot \frac{H}{\beta} \zetaII
    + \frac{1}{\beta^4} 
    + \frac{1}{2H^2\beta^2}\\
    &\le 16d(e^{H/\beta}-1)^2\zetaI + 8(e^{H/\beta} - 1)^2(\frac{H}{\beta} \zetaII + 2),
\end{align*}
where the second inequality is from $2N>1$ and the third inequality is from $(e^{H/\beta}-1)^2\ge (\frac{H}{\beta} + \frac{H^2}{\beta^2})^{2} \ge \frac{H^2}{\beta^2} + \frac{H^4}{\beta^4}\ge \frac{1}{2H^2\beta^2} + \frac{1}{\beta^4}$.
\end{proof}

\begin{lemma}[Concentration of Self-Normalized Processes]
    \label{lem:concentration}
    Let $V:\mathcal{S}\rightarrow [0, H]$ be any fixed function. 
    For any fixed $h\in[H]$, any $0<\delta<1$, all $V\in \mathcal{N}_{\epsilon}(R)$ and all $\beta\in \betaeps$, we have the following holds with probability at least $1-\delta$,
    {\footnotesize
    \begin{align*}
        \lVert \sum_{\tau=1}^{N}\phi(s_h^{\tau},a_h^{\tau})\epsilon_h^{\tau}(\beta, V)\rVert_{\Lambda_h^{-1}}^2> &d\left(e^{H / \beta}-1\right)^2 \log \left(1+\frac{N}{\lambda}\right) +2\left(e^{H / \beta}-1\right)^2 \cdot \log \left(\frac{H\left|\mathcal{N}_\epsilon(R, B, \lambda)\right|\left|\mathcal{N}_\epsilon(\underline{\beta}, \bar{\beta})\right|}{\delta}\right).
    \end{align*}
    }
    \end{lemma}
    \begin{proof}
    Recall the definition of the filtration $\mathcal{F}_h^{\tau}$ and note that $\phi(s_{h}^{\tau},a_{h}^{\tau})\in \mathcal{F}_{h}^{\tau}$. Moreover, for any fixed function $V:\mathcal{S}\rightarrow [0, H]$ and any fixed $\beta\in[\underline{\beta},\overline{\beta}]$, we have $\epsilon_h^{\tau}(\beta, V)\in \mathcal{F}_{h}^{\tau}$ and $\E[\epsilon_h^{\tau}(\beta, V)\lvert \mathcal{F}_h^{\tau}] = \E[(P(\cdot \lvert s_h^{\tau}, a_h^{\tau}) - \mathds{1}(s_{h+1}^{\tau}))^{\top}(e^{(H-V)/\beta}-e^{H/\beta})\lvert \gF_h^{\tau}]= 0$.
    Moreover, as we have $\lvert \epsilon_h(V)\rvert\in [0, e^{H/\beta} - 1]$, for all fixed $h\in [H]$ and all $\tau\in [N]$, $\epsilon_h^{\tau}(\beta, V)$ is mean zero and $(e^{H/\beta}-1)$-sub-Gaussian conditional on $\mathcal{F}_h^{\tau}$.
    
    We invoke Lemma~\ref{lm:self-normalized} with $V=\lambda \cdot I$, $X_t = \phi(s_h^{\tau}, a_h^{\tau})$, $\eta_\tau = \epsilon_h^{\tau}(\beta, V)$ and $R=e^{H/\beta}-1$, we have \begin{align*}
        \mathbb{P}\left (\lVert \sum_{\tau=1}^{N} \phi(x_h^{\tau}, a_h^{\tau})\cdot \epsilon_h(\beta, V)\rVert_{\Lambda_h^{-1}}^2> 2(e^{H/\beta}-1)^2\cdot \log (\frac{\operatorname{det}(\Lambda_h)^{1/2}}{\delta \cdot \operatorname{det}(\lambda \cdot I)^{1/2}})\right)\le \delta,
    \end{align*}
    for any $\delta>0$. Moreover, $\operatorname{det}(\lambda \cdot I) = \lambda^d$, and 
    \begin{align}
    \lVert \Lambda_h\rVert &= \lVert \lambda \cdot I + \sum_{\tau=1}^{N}\phi(s_h^{\tau}, a_h^{\tau})\phi(s_h^{\tau}, a_h^{\tau})^{\top}\rVert \\
    & \le \lambda+\sum_{\tau=1}^{N} \lVert \phi(s_h^\tau, a_h^\tau)\phi(s_h^\tau, a_h^\tau)^{\top}\rVert\\
    &\le \lambda +N,
    \end{align}
    $\operatorname{det}(\Lambda_h)\le (\lambda+N)^d$ for $\Lambda_h$ is a positive-definite matrix.
    Hence we know
    \begin{align*}
        &2(e^{H/\beta}-1)^2\cdot \log \left(\frac{\operatorname{det}(\Lambda_h)^{1/2}}{\delta\cdot \operatorname{det}(\lambda \cdot I)^{1/2}}\right)\\
         &= 2(e^{H/\beta}-1)^2\cdot \frac{d}{2}\log \left(1+\frac{N}{\lambda}\right) + 2(e^{H/\beta}-1)^2\cdot\log(1/\delta)\\
        &\le  d(e^{H/\beta}-1)^2 \log \left(1+\frac{N}{\lambda}\right) + 2(e^{H/\beta}-1)^2\cdot\log(1/\delta), 
    \end{align*}
    which implies
    \begin{align*}
        \sP\left( \lVert \sum_{\tau=1}^{N}\phi(s_h^{\tau},a_h^{\tau})\epsilon_h^{\tau}(\beta, V)\rVert_{\Lambda_h^{-1}}^2> d(e^{H/\beta}-1)^2 \log (1+\frac{N}{\lambda}) + 2(e^{H/\beta}-1)^2\cdot\log(1/\delta)\right)\le \delta.
    \end{align*}
    Finally we know by the union bound that for all $h\in [H]$, the following holds with probability at least $1-\delta$, all $V\in \mathcal{N}_{\epsilon}(R)$ and all $\beta\in \betaeps$,
    {\footnotesize
    \begin{align*}
        \lVert \sum_{\tau=1}^{N}\phi(s_h^{\tau},a_h^{\tau})\epsilon_h^{\tau}(\beta, V)\rVert_{\Lambda_h^{-1}}^2> &d\left(e^{H / \beta}-1\right)^2 \log \left(1+\frac{N}{\lambda}\right) +2\left(e^{H / \beta}-1\right)^2 \cdot \log \left(\frac{H\left|\mathcal{N}_\epsilon(R, B, \lambda)\right|\left|\mathcal{N}_\epsilon(\underline{\beta}, \bar{\beta})\right|}{\delta}\right).
    \end{align*}
    }
\end{proof}

\section{Proof of Theorem~\ref{thm:model-misspecific-insufficient}}
In this section, we mainly prove the Theorem~\ref{thm:model-misspecific-insufficient}. 
By setting the model mis-specification $\xi=0$, we can recover the results in Theorem~\ref{thm:partial-coverage}.

\begin{proof}[Proof of Theorem~\ref{thm:partial-coverage}]
Following the same argument, $\widehat{V}_{h+1}$ depends on $\{(s_{h}^{\tau}, a_{h}^{\tau})\}_{\tau\in [N]}$ via $\{(s_{h^{\prime}}^{\tau}, a_{h^{\prime}}^{\tau})\}_{h^{\prime}>h, \tau\in [N]}$ and thus we cannot directly apply vanilla concentration bounds to control $\lVert \sum_{\tau=1}^N \phi(s_{h}^{\tau}, a_{h}^{\tau})\cdot \eps_h^{\tau}(\beta_{h,i}, \widehat{V}_{h+1})\rVert_{\Lambda_h^{-1}}$. 

To tackle this point, we consider the function family $V_h(R)$. In specific, 
\begin{align*}
    V_h(R) = \{V_h(x;w,\gamma,\Lambda): \mathcal{S}\rightarrow [0, H-h+1] \lvert \lVert w \rVert \le R, \gamma\in[0,B], \Lambda \succeq \lambda \cdot I\},
\end{align*}
and 
\begin{align*}
V_h(x;w, \gamma, \Lambda)&=\max_{a\in \mathcal{A}}\{\max\{\phi(s,a)^{\top}w - \gamma\cdot \sum_{i=1}^d \sqrt{(\phi_i(s,a)\mathds{1}_i)^{\top}\Lambda^{-1}(\phi_i(s,a)\mathds{1}_i)},0\}\}.
\end{align*}

    We continue from the Equation~\ref{eq:decompose-martingale} in Subsection~\ref{subsec:auxiliary} and invoke Lemma~\ref{lem:concentration}, Lemma~\ref{lm:difference-beta} and Lemma~\ref{lm:difference-V} for the term I, II and III respectively and for all $h\in[H]$ and any $\beta\in [\underline{\beta}, \overline{\beta}]$ we have the following holds with probability at least $1-\delta$, ,
\begin{align*}
    & \lVert \sum_{\tau\in[N]} \phi(s_h^{\tau}, a_{h}^{\tau})\epsilon_h^{\tau}(\beta, \widehat{V}_{h+1})\rVert_{\Lambda^{-1}_h}^2 \\
    &\le  4d\left(e^{H / \beta}-1\right)^2 \log \left(1+\frac{N}{\lambda}\right)\\
    & +8\left(e^{H / \beta}-1\right)^2 \cdot \log \left(\frac{H\left|\mathcal{N}_\epsilon(R, B, \lambda)\right|\left|\mathcal{N}_\epsilon(\underline{\beta}, \bar{\beta})\right|}{\delta}\right) \\
    & \quad + \frac{16N^2(H-h)^2\epsilon^2}{\lambda\beta^4}e^{2H/\beta} + \frac{8N^2\epsilon^2}{\lambda\beta^2}e^{2H/\beta}\\
    &\le 4d(e^{H/\beta} - 1)^2\log(1+\frac{N}{\lambda}) + 8(e^{H/\beta} - 1)^2\cdot \log(\frac{H^2}{\epsilon \delta \rho}) + 8 d(e^{H/\beta} - 1)^2\cdot \log(1+\frac{4R}{\epsilon})\\
    &\quad + \frac{16N^2(H-h)^2\epsilon^2}{\lambda\beta^4}e^{2H/\beta} + \frac{8N^2\epsilon^2}{\lambda\beta^2}e^{2H/\beta} + 8(e^{H/\beta} - 1)^2\cdot d^2 \log(1+\frac{8\sqrt{d}B^2}{\lambda \epsilon^2}),
\end{align*}
where the second inequality is from Lemma~\ref{lem:eps-covering-pessimism} and the fact that $\lvert \epsN(\underline{\beta}, \overline{\beta})\rvert \le \frac{H}{\rho\epsilon}$.

By choosing $R=2Hd^{3/2}$, 
$\zeta_2 = \zetaII $, $\zeta_3=\zetaIII$, 
$\lambda=1$, $\gamma = \underline{\beta}(e^{H/\underline{\beta}}-1)(\xi\sqrt{d} + c_1 d\sqrt{\zeta_3})+ c_2(e^{H/\underline{\beta}}-1)\sqrt{H\zeta_2}$, $B=2\gamma$, 
$\epsilon = \frac{2}{4NHe^{H/\beta}}\le \frac{\beta}{2}$, then 
\begin{align*}
    & \lVert \sum_{\tau\in[N]} \phi(s_h^{\tau}, a_{h}^{\tau})\epsilon_h^{\tau}(\beta, \widehat{V}_{h+1})\rVert_{\Lambda^{-1}_h}^2 \\
    &\le 4d(e^{H/\beta} - 1)^2\cdot \log(2N) 
    + 8(e^{H/\beta} - 1)^2\cdot \frac{H}{\beta} \zetaII
    + \frac{1}{\beta^4} 
    + \frac{1}{2H^2\beta^2}\\
    & \quad 
    + 8d(e^{H/\beta} - 1)^2 \cdot \log(1+8NH^2e^{\frac{H}{\underline{\beta}}}) + 8d^2(e^{H/\beta} - 1)^2\cdot  \log(1+32(c^{\prime})^2N^2H^3d^{5/2}\gamma e^{2H/\underline{\beta}}) 
    \\
    &\le 16d(e^{H/\beta}-1)^2 \log(2N + 8NH^2e^{H/\underline{\beta}})  
    + \frac{1}{\beta^4} 
    + \frac{1}{2H^2\beta^2}\\
    &\quad + 8(e^{H/\beta} - 1)^2\cdot \frac{H}{\beta} \zetaII
    + 8d^2(e^{H/\beta} - 1)^2\cdot \log(1+32(c^{\prime})^2N^2H^3d^{5/2}\gamma e^{2H/\underline{\beta}}) \\
    & \le 16d(e^{H/\beta}-1)^2\zetaI + 8(e^{H/\beta} - 1)^2(\frac{H}{\beta} \zetaII + 2) \\
    &\quad + 8d^2(e^{H/\beta} - 1)^2\cdot \log(1+32(c^{\prime})^2N^2H^3d^{5/2}\gamma e^{2H/\underline{\beta}})\\
    &\le 32d^2(e^{H/\beta} - 1)^2\cdot\log(2N+32(c^{\prime})^2N^2H^3d^{5/2}\gamma e^{2H/\underline{\beta}}) +  8(e^{H/\beta} - 1)^2\cdot (\frac{H}{\beta} \zetaII
    +2)\\
    &\le 64d^2(e^{H/\beta} - 1)^2\cdot\log(2N+32(c^{\prime})^2N^2H^3d^{5/2}\gamma e^{2H/\underline{\beta}}) +  8(e^{H/\beta} - 1)^2\cdot \frac{H}{\beta} \zetaII
\end{align*}

Thus we have 
\begin{align*}
    &\max_{i\in[d],\beta_{h,i}\in[\underline{\beta},\overline{\beta}]} 
    \beta_{h,i}\lVert \sum_{\tau=1}^N \phi(s_{h}^{\tau}, a_{h}^{\tau})\cdot \eps_h^{\tau}(\beta_{h,i}, \widehat{V}_{h+1})\rVert_{\Lambda_h^{-1}}\\
    &\le \max_{i\in[d],\beta_{h,i}\in[\underline{\beta},\overline{\beta}]} 
    8d \beta_{h,i} (e^{H/\beta_{h,i}} - 1)\cdot\sqrt{\zetaIII}\\
    &\quad +  \max_{i\in[d],\beta_{h,i}\in[\underline{\beta},\overline{\beta}]} 2\sqrt{2}\beta_{h,i} (e^{H/\beta_{h,i}} - 1)\cdot \sqrt{\frac{H}{\beta_{h,i}} \zetaII}\\
    &  \le 
    \max_{i\in[d],\beta_{h,i}\in[\underline{\beta},\overline{\beta}]} 
    8 d \beta_{h,i} (e^{H/\beta_{h,i}} - 1)\cdot\sqrt{\zetaIII}\\
    &\quad 
    +  \max_{i\in[d],\beta_{h,i}\in[\underline{\beta},\overline{\beta}]} 2\sqrt{2\beta_{h,i}}(e^{H/\beta_{h,i}} - 1)\cdot \sqrt{H\zetaII}
    \\
    &\le 8 d\underline{\beta} (e^{H/\underline{\beta}}-1) \sqrt{\zeta_3} + 2\sqrt{2} \sqrt{\underline{\beta}}(e^{H/\underline{\beta}}-1)\sqrt{H\zeta_2} 
,
\end{align*}

holds for probability at least $1-\delta$ for all $h\in[H]$ for some constant $c>1$ and $\zeta_2=\zetaII$ and $\zeta_3 = \zetaIII$.
Thus we have 
\begin{equation}
\label{eq:error-w-pessimism}
\begin{aligned}
    &\lvert \phi(s,a)^{\top}(w_h-\widehat{w}_h)\rvert\\
    & \le 
    \max_{i\in[d], \beta_{h,i}\in[\underline{\beta},\overline{\beta}]} 
    \beta_{h,i}(e^{H/\beta_{h,i}}-1)(\xi\sqrt{d}  + 8 d\sqrt{\zeta_3} + \sqrt{d})\sum_{i=1}^d \lVert \phi_i(s,a)\mathds{1}_i\rVert_{\Lambda_h^{-1}} \\
    & \quad + 2\sqrt{2}\sqrt{\beta_{h,i}}(e^{H/\beta_{h,i}}-1)\sqrt{H\zeta_2}\sum_{i=1}^d \lVert \phi_i(s,a)\mathds{1}_i\rVert_{\Lambda_h^{-1}}\\
    &\le 
    \underline{\beta}(e^{H/\underline{\beta}}-1)
    (\xi\sqrt{d}  + 9 d\sqrt{\zeta_3} )
    \sum_{i=1}^d \lVert \phi_i(s,a)\mathds{1}_i\rVert_{\Lambda_h^{-1}} 
    + 2\sqrt{2} \sqrt{\underline{\beta}}(e^{H/\underline{\beta}}-1)\sqrt{H\zeta_2}\sum_{i=1}^d \lVert \phi_i(s,a)\mathds{1}_i\rVert_{\Lambda_h^{-1}},
\end{aligned}
\end{equation}
where the last inequality is by noticing that $f(\beta) = \beta (e^{H/\beta}-1)$ and $g(\beta) = \sqrt{\beta}(e^{H/\beta}-1)$ are both monotonically decreasing with $\beta>0$.

Plug Equation~\ref{eq:error-w-pessimism} and \ref{eq:error-theta} into Equation~\ref{eq:Bellman-error} to finally upper bound the Bellman error with the choice $\lambda = 1$,
\begin{align*}
    \lvert \iota_h(s,a)\rvert& \le \sqrt{d} \sum_{i=1}^d \lVert \phi_i(s,a)\mathds{1}_i\rVert_{\Lambda_h^{-1}} + \underline{\beta}(e^{H/\underline{\beta}}-1)
    (\xi\sqrt{d}  + 9 d\sqrt{\zeta_3})
    \sum_{i=1}^d \lVert \phi_i(s,a)\mathds{1}_i\rVert_{\Lambda_h^{-1}} \\
    &\quad + 2\sqrt{2} \sqrt{\underline{\beta}}(e^{H/\underline{\beta}}-1)\sum_{i=1}^d \lVert \phi_i(s,a)\mathds{1}_i\rVert_{\Lambda_h^{-1}} + (H-h)\xi\\
    &\le \underline{\beta}(e^{H/\underline{\beta}}-1)(\xi\sqrt{d} + 10 d\sqrt{\zeta_3})\sum_{i=1}^d \lVert \phi_i(s,a)\mathds{1}_i\rVert_{\Lambda_h^{-1}}+ 2\sqrt{2} \sqrt{\underline{\beta}}(e^{H/\underline{\beta}}-1)\sqrt{H\zeta_2}\sum_{i=1}^d \lVert \phi_i(s,a)\mathds{1}_i\rVert_{\Lambda_h^{-1}}\\
    &\quad + (H-h)\xi,
\end{align*}
where the last inequality is from the fact that $f(\beta) = \beta(e^{H/\beta}-1)\ge H\ge 1$ and and $\sqrt{d\zeta_3}>1$ as $N\ge e/2$.





Using the similar steps in the proof of Corollar 4.5 in \cite{jin2021pessimism}, we know that 
\begin{align*}
    \operatorname{SubOpt}(\widehat{\pi};\gP) &\le c^{\prime} \sum_{h=1}^H \mathbb{E}[\sum_{i=1}^d\sqrt{(\phi_i(s,a)\mathds{1}_i)^{\top} \Lambda_h^{-1}(\phi_i(s,a)\mathds{1}_i)}] + H(H-1)\xi/2\\
    &\le c^{\prime} \sum_{h=1}^H \mathbb{E}[\sum_{i=1}^d\sqrt{\operatorname{Tr}(\Lambda_h^{-1} (\phi_i(s,a)\mathds{1}_i)(\phi_i(s,a)\mathds{1}_i)^{\top})}]+ H(H-1)\xi/2\\
    &\le c^{\prime} \sum_{h=1}^H \mathbb{E}[\sum_{i=1}^d\sqrt{\frac{\lambda_{h,i,j}}{1+c^{\dagger}\cdot N\cdot \lambda_{h,i,j}}}]+ H(H-1)\xi/2\\
    &\le c^{\prime} \sum_{h=1}^H \mathbb{E}[\sum_{i=1}^d\sqrt{\frac{1}{1+c^{\dagger}\cdot N}}]+ H(H-1)\xi/2\\
    &\le c^{\prime} \cdot d^{1/2} H/\sqrt{N} + H(H-1)\xi/2,
\end{align*}
where $c^{\prime}\coloneqq c_1 \underline{\beta}(e^{H/\underline{\beta}}-1)(\xi d +  d^{3/2}\sqrt{\zeta_3})H/ + c_2\sqrt{\underline{\beta}}(e^{H/\underline{\beta}}-1)d^{1/2}H^{3/2}\sqrt{\zeta_2}$ and $c_1, c_2$ are some absolute constants only depend on $c^{\dagger}$.

In conclusions, we have probability at least $1-\delta$, 
\begin{align*}
        \operatorname{SubOpt}(\widehat{\pi};\gP)\le & c_1 \underline{\beta}(e^{H/\underline{\beta}}-1)(\xi d +  d^{3/2}\sqrt{\zeta_3})H/N^{1/2} + c_2\sqrt{\underline{\beta}}(e^{H/\underline{\beta}}-1)d^{1/2}H^{3/2}\sqrt{\zeta_2}/N^{1/2}\\
        & \quad + H(H-1)\xi/2.
\end{align*}
\end{proof}

\section{Auxiliary Lemma}
Before we proceed our analysis, we need the following lemmas.

\begin{fact}
\label{fa:exp}
For $x,y\ge 0$ and $b>0$, we have $\lvert e^{-by} - e^{-bx}\rvert\le b\lvert x-y\rvert$.
\end{fact}

\begin{fact}
\label{fa:exp-2}
For $x,y\ge 0$ and $b>0$, we have $b\lvert x-y\rvert\le \lvert e^{bx}-e^{by}\rvert$.
\end{fact}

\begin{lemma}[$\epsilon$-Covering Number \cite{jin2020provably}]
    \label{lem:eps-covering}
        Let $\mathcal{V}$ denote a class of functions mapping from $\mathcal{S}$ to $\mathbb{R}$ with following parametric form
        $$
        V(s)=\max _a w^{\top} \boldsymbol{\phi}(s, a),
        $$
        where the parameters $w$ satisfy $\|w\| \leq L, \beta \in[0, B]$ Assume $\|\phi(x, a)\| \leq 1$ for all $(x, a)$ pairs, and let $\mathcal{N}_{\varepsilon}$ be the $\varepsilon$-covering number of $\mathcal{V}$ with respect to the distance $\operatorname{dist}\left(V, V^{\prime}\right)=\sup _x\left|V(x)-V^{\prime}(x)\right|$. Then
        $$
        \log \lvert \mathcal{N}_{\epsilon}(R, B, \lambda)\rvert\le ds \log(1+4R/\epsilon).
        $$
\end{lemma}

\begin{lemma}
    \label{lem:eps-covering-pessimism}
Let $\mathcal{V}$ denote a class of functions mapping from $\mathcal{S}$ to $\mathbb{R}$ with following parametric form
$$
V(s)=\max _a w^{\top} \boldsymbol{\phi}(s, a)+\beta \sum_{i=1}^d \sqrt{(\phi_i(s, a)\mathds{1}_i)^{\top} \Lambda^{-1} (\phi_i(s, a)\mathds{1}_i)},
$$
where the parameters $(w, \beta, \Lambda)$ satisfy $\|w\| \leq L, \beta \in[0, B]$ and the minimum eigenvalue satisfies $\lambda_{\min }(\Lambda) \geq \lambda$. Assume $\|\phi(x, a)\| \leq 1$ for all $(x, a)$ pairs, and let $\mathcal{N}_{\varepsilon}$ be the $\varepsilon$-covering number of $\mathcal{V}$ with respect to the distance $\operatorname{dist}\left(V, V^{\prime}\right)=\sup _x\left|V(x)-V^{\prime}(x)\right|$. Then
$$
\log \mathcal{N}_{\varepsilon} \leq d \log (1+4 L / \varepsilon)+d^2 \log \left[1+8 d^{1 / 2} B^2 /\left(\lambda \varepsilon^2\right)\right].
$$
\end{lemma}
\begin{proof}
Equivalently, we can reparametrize the function class $\mathcal{V}$ by let $A=\beta^2 \Lambda^{-1}$, so we have
$$
V(s)=\max_a w^{\top} \phi(s, a)+\sum_{i=1}^d \sqrt{(\phi_i(s,a)\mathds{1}_i)^{\top} A (\phi_i(s,a)\mathds{1}_i)},
$$
for $\|w\| \leq L$ and $\|A\| \leq B^2 \lambda^{-1}$. For any two functions $V_1, V_2 \in \mathcal{V}$, let them take the form in Eq. (27) with parameters $\left(w_1, A_1\right)$ and $\left(w_2, A_2\right)$, respectively. Then, since both $\min \{\cdot, H\}$ and $\max _a$ are contraction maps, we have
$$
\begin{aligned}
\operatorname{dist}\left(V_1, V_2\right) & \leq \sup _{x, a}\left|\left[w_1^{\top} \phi(x, a)+\sum_{i=1}^d \sqrt{(\phi_i(s,a)\mathds{1}_i)^{\top} A_2 (\phi_i(s,a)\mathds{1}_i)}\right]-\left[w_2^{\top} \phi(x, a)+\sum_{i=1}^d\sqrt{(\phi_i(s,a)\mathds{1}_i)^{\top} A_1 (\phi_i(s,a)\mathds{1}_i)}\right]\right| \\
& \leq \sup _{\phi:\|\phi\| \leq 1}\left|\left[w_1^{\top} \phi+\sum_{i=1}^d \sqrt{(\phi_i(s,a)\mathds{1}_i)^{\top} A_1 (\phi_i(s,a)\mathds{1}_i)}\right]-\left[w_2^{\top} \phi+\sum_{i=1}^d \sqrt{(\phi_i(s,a)\mathds{1}_i)^{\top} A_2 (\phi_i(s,a)\mathds{1}_i)}\right]\right| \\
& \leq \sup _{\phi:\|\phi\| \leq 1}\left|\left(w_1-w_2\right)^{\top} \phi\right|+\sup _{\phi:\|\phi\| \leq 1} \sum_{i=1}^d\sqrt{\left|(\phi_i(s,a)\mathds{1}_i)^{\top}\left(A_1-A_2\right) (\phi_i(s,a)\mathds{1}_i)\right|} \\
& =\left\|w_1-w_2\right\|+\sqrt{\left\|A_1-A_2\right\|}\sup _{\phi:\|\phi\| \leq 1} \sum_{i=1}^d \lVert \phi_i(s,a)\mathds{1}_i \rVert\\
& \leq\left\|w_1-w_2\right\|+\sqrt{\left\|A_1-A_2\right\|_F},
\end{aligned}
$$
where the second last inequality follows from the fact that $|\sqrt{x}-\sqrt{y}| \leq \sqrt{|x-y|}$ holds for any $x, y \geq 0$. 
For matrices, $\|\cdot\|$ and $\|\cdot\|_F$ denote the matrix operator norm and Frobenius norm respectively.

Let $\mathcal{C}_{w}$ be an $\varepsilon / 2$-cover of $\left\{w \in \mathbb{R}^d \mid\|w\| \leq L\right\}$ with respect to the 2-norm, and $\mathcal{C}_{A}$ be an $\varepsilon^2 / 4$-cover of $\left\{A \in \mathbb{R}^{d \times d} \mid\|A\|_F \leq d^{1 / 2} B^2 \lambda^{-1}\right\}$ with respect to the Frobenius norm. By Lemma D.5, we know:
$$
\left|\mathcal{C}_{w}\right| \leq(1+4 L / \varepsilon)^d, \quad\left|\mathcal{C}_{A}\right| \leq\left[1+8 d^{1 / 2} B^2 /\left(\lambda \varepsilon^2\right)\right]^{d^2} .
$$
By Eq. (28), for any $V_1 \in \mathcal{V}$, there exists $w_2 \in \mathcal{C}_{w}$ and $A_2 \in \mathcal{C}_{A}$ such that $V_2$ parametrized by $\left(w_2, A_2\right)$ satisfies $\operatorname{dist}\left(V_1, V_2\right) \leq \varepsilon$. Hence, it holds that $\mathcal{N}_{\varepsilon} \leq\left|\mathcal{C}_{w}\right| \cdot\left|\mathcal{C}_{A}\right|$, which gives:
$$
\log \mathcal{N}_{\varepsilon} \leq \log \left|\mathcal{C}_{w}\right|+\log \left|\mathcal{C}_{A}\right| \leq d \log (1+4 L / \varepsilon)+d^2 \log \left[1+8 d^{1 / 2} B^2 /\left(\lambda \varepsilon^2\right)\right]
$$
This concludes the proof.
\end{proof}

\begin{lemma}[Self-Normalized Bound for Vector-Valued Martingales \citep{abbasi2011improved}]
\label{lm:self-normalized}
Let $\{\mathcal{F}_t\}_{t=0}^{\infty}$ be a filtration. Let $\{\eta_t\}_{t=1}^{\infty}$ be a real-valued stochastic process such that $\eta_t$ is $\mathcal{F}_t$-measurable and $\eta_t$ is conditionally $R$-sub-gaussian for some $R\ge 0$, i.e., 
\begin{align*}
    \forall \lambda\in \R, \E[e^{\lambda \eta_t}\lvert \mathcal{F}_{t-1}]\le e^{\frac{\lambda^2 R^2}{2}}.
\end{align*}
Let $\{X_t\}_{t=1}^{\infty}$ be an $\R^d$-valued stochstic process such that $X_t$ is $\mathcal{F}_{t-1}$-measurable. Assume that $V$ is a $d\times d$ positive definite matrix. For any $t\ge 0$, define 
\begin{align*}
    V_t = V+\sum_{s=1}^\top X_sX_s^{\top}, \quad S_t = \sum_{s=1}^\top \eta_s X_s.
\end{align*}
Then for any $\delta>0$, with probability at least $1-\delta$, for all $t\ge 0$, 
\begin{align*}
    \lVert S_t\rVert_{V_t^{-1}}^2\le 2R^2 \log(\frac{\det(V_t)^{1/2}\det(V)^{-1/2}}{\delta}).
\end{align*}
\end{lemma}

\begin{lemma}
    \label{lm:exp-beta-difference}
    For any $0\le x\le H$, $\lvert \beta^{\prime} - \beta \rvert\le \epsilon$ for some $\epsilon>0$, we have 
    \begin{align*}
        \lvert e^{(H-x)/\beta^\prime} - e^{(H-x)/\underline{\beta}}\rvert \le \expbeta.
    \end{align*}
    \end{lemma}
    \begin{proof}
    We denote $f(z) = e^{(H-x)/z}$. 
    Then 
    \begin{align*}
        \lvert f^{\prime}(z)\rvert &= \lvert e^{(H-x)/z}\cdot \frac{(H-x)}{z^2}\rvert\\
        &\le e^{H/z}\cdot \frac{H}{z^2}\\
        &\le e^{2H/\beta}\cdot \frac{4H}{\beta^2},
    \end{align*}
    where the last inequality is from the fact that $z\ge \beta-\epsilon\ge \frac{\beta}{2}$.
    Thus by the mean value theorem we know 
    \begin{align*}
        \lvert e^{(H-x)/\beta^\prime} - e^{(H-x)/\beta}\rvert \le \expbeta.
    \end{align*}
    \end{proof}
    
    \begin{lemma}
    For any $\beta\in[\underline{\beta}, \overline{\beta}]$, and any $\epsilon$-net over $[\underline{\beta}, \overline{\beta}]$, i.e., $\mathcal{N}_{\epsilon}(\beta)$ for some $\epsilon>0$, we have
    \label{lm:eps-beta-difference}
    \begin{align*}
        \lvert \epsilon(\beta, V) - \epsilon(\epsN(\beta), V)\rvert \le \epsbeta.
    \end{align*}
    \end{lemma}
    \begin{proof}
    \begin{align*}
        & \lvert \epsilon(\beta, V) - \epsilon(\epsN(\beta), V)\rvert\\
        &= \lvert \E_{s^{\prime}}[e^{(H-V(s^{\prime}))/\beta}] - e^{(H-V(s_{h+1}^{\tau}))/\beta} - (\E_{s^{\prime}}[e^{(H-V(s^{\prime}))/\epsN(\beta)}] - e^{(H-V(s_{h+1}^{\tau}))/\epsN(\beta)})\rvert \\
        &= \lvert \E_{s^{\prime}}[e^{(H-V(s^{\prime}))/\beta} - e^{(H-V(s^{\prime}))/\epsN(\beta)}]\rvert + \lvert (e^{(H-V(s_{h+1}^{\tau}))/\beta} - e^{(H-V(s_{h+1}^{\tau}))/\epsN(\beta)})\rvert \\
        &\le \epsbeta,
    \end{align*}
    where the last inequality is form Lemma~\ref{lm:exp-beta-difference}.
    \end{proof}
    
    \begin{lemma}
    \label{lm:eps-v-difference}
    For any $V^{\dagger}$ and $V^{\ddagger}:\gS\rightarrow [0,H]$, and $\lVert V^{\dagger} - V^{\ddagger}\rVert_{\infty}\le \epsilon$, for some $\epsilon>0$, we have 
    \begin{align*}
         \lvert \epsilon(\beta, V^{\dagger}) - \epsilon(\beta, V^{\ddagger})\rvert\le \epsVValue.
    \end{align*}
    \end{lemma}
    \begin{proof}
    \begin{align*}
        &\lvert \epsilon(\beta, V^{\dagger}) - \epsilon(\beta, V^{\ddagger})\rvert\\
        &= \lvert \E_{s^{\prime}}[e^{(H-V^{\dagger}(s^{\prime}))/\beta}] - e^{(H-V^{\dagger}(s^{\prime}))/\beta} - (\E_{s^{\prime}}[e^{(H-V^{\ddagger}(s^{\prime}))/\beta}] - e^{(H-V^{\ddagger}(s^{\prime}))/\beta})\rvert \\
        &= \lvert \E_{s^{\prime}}[e^{(H-V^{\dagger}(s^{\prime}))/\beta} - e^{(H-V^{\ddagger}(s^{\prime}))/\beta}]\rvert +  \lvert (e^{(H-V^{\dagger}(s^{\prime}))/\beta} - e^{(H-V^{\ddagger}(s^{\prime}))/\beta})\vert \\
        &= e^{H}\lvert \E_{s^{\prime}}[e^{(-V^{\dagger}(s^{\prime}))/\beta} - e^{(-V^{\ddagger}(s^{\prime}))/\beta}\rvert + \lvert(e^{(-V^{\dagger}(s^{\prime}))/\beta} - e^{(-V^{\ddagger}(s^{\prime}))/\beta})\vert \\
        &\le \epsVValue,
    \end{align*}
    where the last inequality is form Fact~\ref{fa:exp}.
    \end{proof}
        
\begin{lemma}
    \label{lm:difference-beta}
    \begin{align*}
        \lVert \sum_{\tau\in[N]} \left( \phi(s_h^{\tau}, a_{h}^{\tau})\left(\epsilon_h^{\tau}(\beta, V) - \epsilon_h^{\tau}(N_\epsilon(\beta), V)\right) \right) \rVert_{\Lambda^{-1}_h}^2\le \epsbetas.
    \end{align*}
\end{lemma}
\begin{proof}
    \begin{align*}
        &\lVert \sum_{\tau\in[N]} \left( \phi(s_h^{\tau}, a_{h}^{\tau})\left(\epsilon_h^{\tau}(\beta, V) - \epsilon_h^{\tau}(N_\epsilon(\beta), V)\right) \right) \rVert_{\Lambda^{-1}_h}^2\\
    &\le \sum_{\tau,\tau^{\prime}}^N \phi(s_{h}^{\tau}, a_{h}^{\tau})^{\top}\Lambda_h^{-1}\phi(s_{h}^{\tau}, a_{h}^{\tau})\cdot(\epsilon_h^{\tau}(\beta, V)-\epsilon_h^{\tau}(N_\epsilon(\beta), V))\cdot (\epsilon_h^{\tau^{\prime}}(\beta, V)-\epsilon_h^{\tau^{\prime}}(N_{\epsilon}(\beta), V)\\
    &\le \frac{4(H-h)^2\epsilon^2}{\underline{\beta}^4}e^{2H/\beta}\sum_{\tau,\tau^{\prime}}^N \lvert \phi(s_{h}^{\tau}, a_{h}^{\tau})^{\top}\Lambda_h^{-1}\phi(s_{h}^{\tau}, a_{h}^{\tau}) \rvert \\
    &\le \frac{4(H-h)^2\epsilon^2}{\underline{\beta}^4}e^{2H/\beta}\sum_{\tau,\tau^{\prime}}^N \lVert \phi(s_{h}^{\tau}, a_{h}^{\tau})\rVert^2 \lVert \Lambda_h^{-1}\rVert\\
    &\le \epsbetas,
    \end{align*}
    where the first inequality is from Lemma~\ref{lm:eps-beta-difference}.
\end{proof}

\begin{lemma}
    \label{lm:difference-V}
    \begin{align*}
        \lVert \sum_{\tau\in[N]} \phi(s_h^{\tau}, a_{h}^{\tau})(\epsilon_h^{\tau}(\beta, \widehat{V}_{h+1}) - \epsilon_h^{\tau}(\beta, V^{\dagger}_{h+1})) \rVert_{\Lambda^{-1}_h}^2\le \epsVs.
    \end{align*}
\end{lemma}
\begin{proof}
    \begin{align*}
        &\lVert \sum_{\tau\in[N]} \phi(s_h^{\tau}, a_{h}^{\tau})(\epsilon_h^{\tau}(\beta, \widehat{V}_{h+1}) - \epsilon_h^{\tau}(\beta, V^{\dagger}_{h+1})) \rVert_{\Lambda^{-1}_h}^2\\
        &\le \sum_{\tau,\tau^{\prime}}^N \phi(s_{h}^{\tau}, a_{h}^{\tau})^{\top}\Lambda_h^{-1}\phi(s_{h}^{\tau}, a_{h}^{\tau})\cdot(\epsilon_h^{\tau}(\beta, \widehat{V}_{h+1})-\epsilon_h^{\tau}(\beta, V^{\dagger}_{h+1}))\cdot (\epsilon_h^{\tau^{\prime}}(\beta, \widehat{V}_{h+1})-\epsilon_h^{\tau^{\prime}}(\beta, V^{\dagger}_{h+1})\\
        &\le \frac{4\epsilon^2}{\underline{\beta}^2}e^{2H/\underline{\beta}} \sum_{\tau,\tau^{\prime}}^N \lvert \phi(s_h^{\tau}, a_h^{\tau})^{\top}\Lambda_h^{-1}\phi(s_h^{\tau^{\prime}}, a_h^{\tau^{\prime}})\rvert\\
        &\le \frac{4\epsilon^2}{\underline{\beta}^2}e^{2H/\underline{\beta}} \sum_{\tau,\tau^{\prime}}^N \lVert \phi(s_h^{\tau}, a_h^{\tau})\rVert \lvert \phi(s_h^{\tau^{\prime}}, a_h^{\tau^{\prime}})\rVert \rvert \lVert \Lambda_h^{-1}\rVert\\
        &\le \epsVs,
    \end{align*}
\end{proof}
where the first inequality is from Lemma~\ref{lm:eps-v-difference} and the last inequality is from the fact that $\lVert x\rVert_2\le \lVert x\rVert_1$ for any $x\in \R^d$ and Assumption~\ref{as:d-rec} that $\lVert \phi(s,a)\rVert_1 = 1$ and also the fact that $\lVert \Lambda_h^{-1}\rVert\le \frac{1}{\lambda}$.

\section{Reproduction of RPVI}
\label{app:RPVI}
In this part, we introduce the implementation of RPVI~\cite{tamar2013scaling} in our setting. Since the original RPVI focuses on the online setting with general uncertainty set, we instantiate RPVI with the episode MDP and KL-divergence as the uncertainty measure. Similar to our method, we incorporate Lemma~\ref{lem:huKL} to robust problem for each $(s,a)$: 
\begin{equation} \label{equ:g_sa}
    \sigma_{sa}(V) = \sup _{\beta \in[0, \infty)}\left\{-\beta \log \left(\mathbb{E}_{P(\cdot \mid s,a)}e^{-V(\cdot) / \beta}\right)-\rho \beta\right\}.
\end{equation}

In the offline setting, the main challenge is to estimate the $\mathbb{E}_{P(\cdot \mid s,a)}e^{-V(\cdot) / \beta}$ from data. Since the ordinary least squares (OLS) has the close form solution, we can estimate Equation~\ref{equ:g_sa} with  
\begin{equation}
    \hat{\sigma}_{sa}(V) = \sup _{\beta \in[0, \infty)}\left\{-\beta \log \left(
    \phi(s,a)^\top 
    \Lambda_h^{-1}
    \left(\sum_{\tau=1}^K 
    \phi\left(s_h^\tau, a_h^\tau\right) \cdot \left(e^{-V(s_{h+1}^\tau) / \beta}\right) \right) 
    \right)-\rho \beta\right\}.
\end{equation}
Plugging it into the template of RPVI, we have the algorithm in Algorithm~\ref{alg:rpvi}.

\begin{algorithm}[ht]
    \caption{RPVI with KL-divergence}
    \label{alg:rpvi}
   \begin{algorithmic}[1]
      \STATE {\bfseries Input:} $\underline{\beta}$, $ \mathcal{D}=\left\{\left(s_{h}^{\tau}, a_{h}^{\tau}, r_{h}^{\tau}\right)\right\}_{\tau, h=1}^{N, H}$.
      \STATE {\bfseries Init:} $\widehat{V}_{H} = 0$.
      \FOR {step $h=H$ {\bfseries to} $1$}
      \STATE $\Lambda_{h}
      \leftarrow \sum_{\tau=1}^{N} \phi\left(s_{h}^{\tau}, a_{h}^{\tau}\right) \phi\left(s_{h}^{\tau}, a_{h}^{\tau}\right)^{\top}+\lambda I$
      \IF {$h=H$}
      \STATE $\widehat{w}_H \gets \Lambda_H^{-1} \left[\sum_{\tau=1}^{N} \phi(s_h^\tau, a_h^\tau) r_h^\tau \right] $
      \ELSE 
    \STATE $\widehat{w}_h \gets \Lambda_h^{-1}\left[\sum_{\tau=1}^N \phi\left(s_h^\tau, a_h^\tau\right) 
    \left(
    r_h^\tau + \hat{\sigma}_{sa}(\hat{V}_{h+1})
    \right)
    \right]$
       \ENDIF
      \STATE $\widehat{Q}_h(\cdot,\cdot) \gets \phi(\cdot,\cdot)^\top \widehat{w}_h$
     \STATE $\widehat{\pi}_{h}(\cdot \mid \cdot) \leftarrow \arg \max _{\pi_{h}}\langle\widehat{Q}_{h}(\cdot, \cdot), \pi_{h}(\cdot \mid \cdot)\rangle_{\mathcal{A}}$
     \STATE $\widehat{V}_{h}(\cdot) \leftarrow\langle\widehat{Q}_{h}(\cdot, \cdot), \widehat{\pi}_{h}(\cdot \mid \cdot)\rangle_{\mathcal{A}}$
      \ENDFOR
   \end{algorithmic}
\end{algorithm}

\begin{figure*}[htbp]
\centering
\subfigure[Average Return]{
	\label{fig.tamar.1}
	\includegraphics[width=0.31\textwidth]{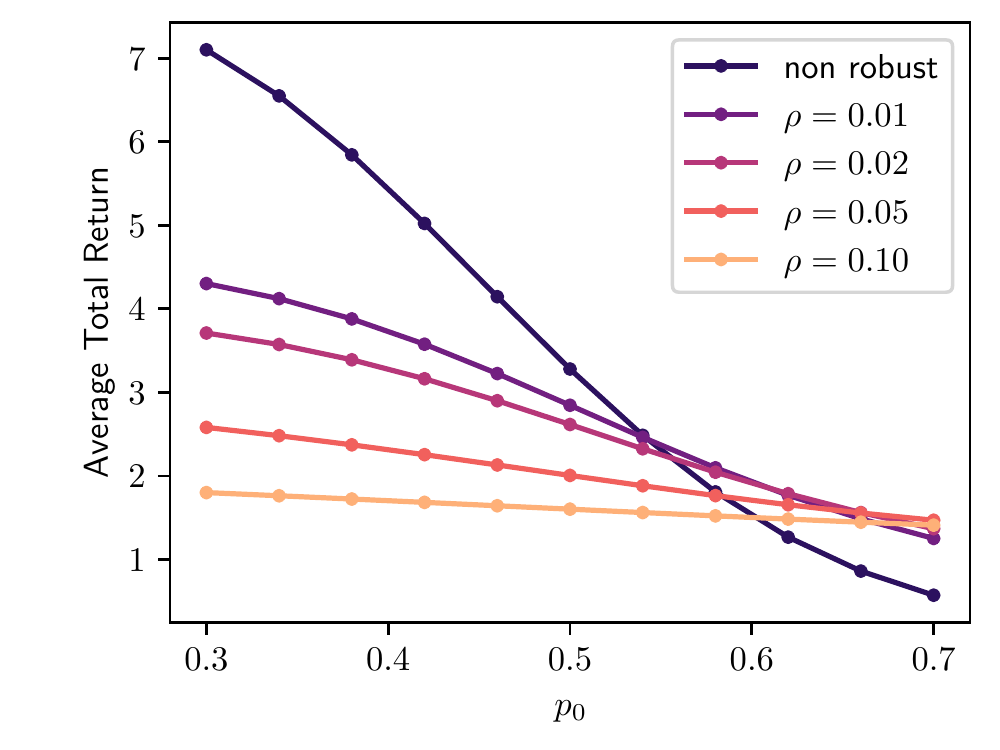}} 
\hfill
\subfigure[$\| \widehat{V}_1 - V^*_1\|$]{
\label{fig.tamar.2}
\includegraphics[width=0.31\textwidth]{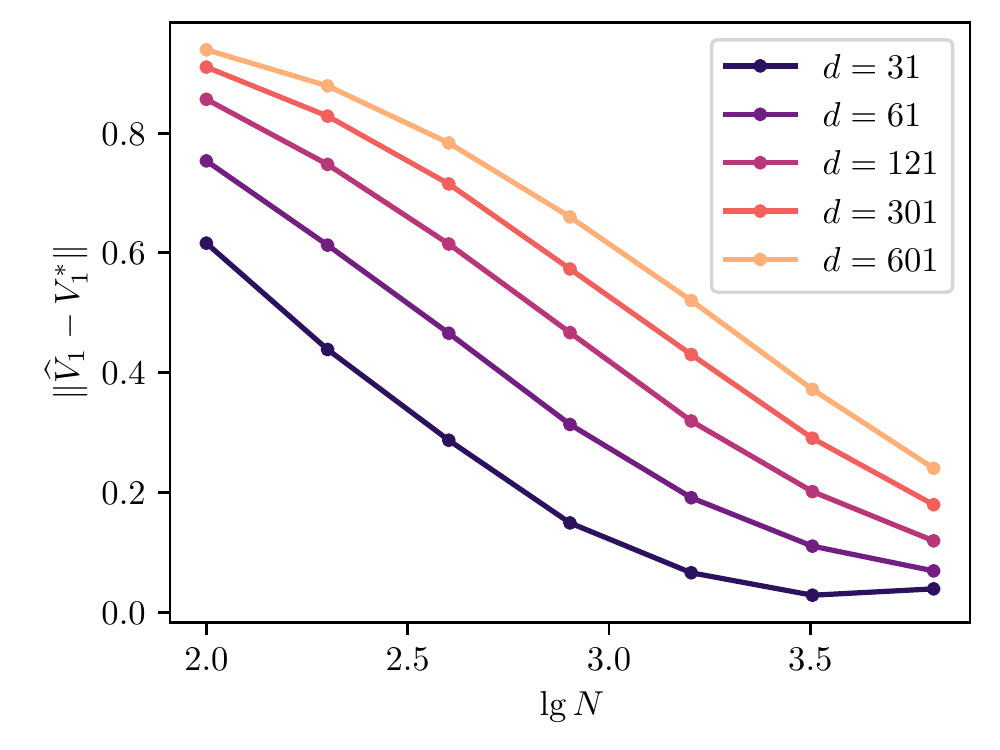}} \hfill
\subfigure[Execution time]{
\label{fig.tamar.3}
\includegraphics[width=0.31\textwidth]{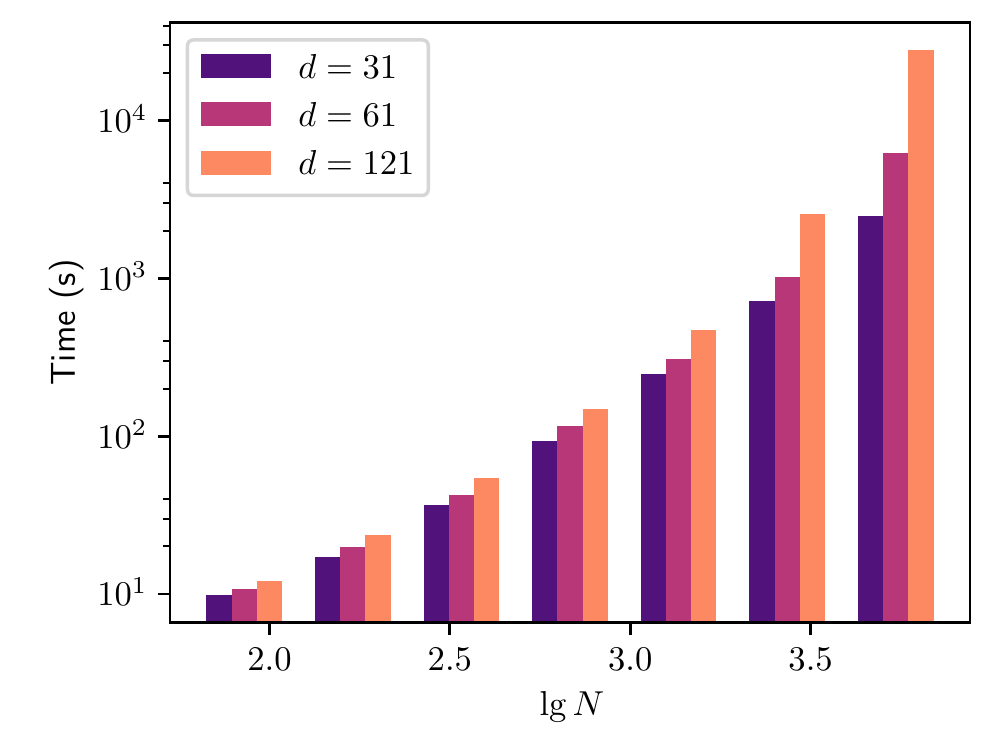}}
\hfill
\caption{Results of RPVI in American Option Experiment. (a) Average total return of different KL radius $\rho$ in the perturbed environments ($N = 1000$, $d=61$). (b) Estimation error with different linear function dimension $d$'s and the sizes of dataset $N$'s ($\rho=0.01$).  (c) Execution time for different $d$'s, \textbf{which is shown in log space}.}
\label{fig.tamar}
\end{figure*}

The results of American Option Experiment are shown in Figure~\ref{fig.tamar}. In the same experimental setup (with the same $\rho$ and $N$), our algorithm DRVI-L has higher average return values, more accurate value estimates, and faster computing time than PRVI.

\end{document}